\documentclass[12pt, fullpage]{article}
\usepackage[utf8]{inputenc}
\usepackage[backend=biber, style=numeric-comp, sorting=none, maxbibnames=20, safeinputenc]{biblatex}
\usepackage{siunitx}
\usepackage{graphicx}
\graphicspath{{images/}}
\usepackage{geometry}
\usepackage{amsmath}
\usepackage{listings}
\usepackage{setspace}
\usepackage[
  colorlinks=true,     
  linkcolor=blue,      
  citecolor=teal,      
  urlcolor=magenta     
]{hyperref}

\usepackage{caption}
\usepackage{subcaption}
\usepackage{xcolor}
\usepackage{indentfirst}
\definecolor{codegreen}{rgb}{0,0.6,0}
\definecolor{codegray}{rgb}{0.5,0.5,0.5}
\definecolor{codeorange}{rgb}{1,0.5,0}
\definecolor{backcolor}{rgb}{0.95,0.95,0.95}

\lstdefinestyle{thesis_style}{
    backgroundcolor=\color{backcolor},   
    commentstyle=\color{codegreen},
    keywordstyle=\color{codeorange},
    numberstyle=\tiny\color{codegray},
    stringstyle=\color{red},
    basicstyle=\ttfamily\footnotesize,
    breakatwhitespace=false,         
    breaklines=true,                 
    captionpos=b,                    
    keepspaces=false,                 
    numbers=left,                    
    numbersep=5pt,                  
    showspaces=false,                
    showstringspaces=false,
    showtabs=false,                  
    tabsize=2,
    xleftmargin=10pt
}

\usepackage{float}

\DeclareFieldFormat*{title}{#1}

\begin{document}
\doublespacing

\pagenumbering{roman}
\thispagestyle{empty}

\begin{titlepage}
\centering

{\LARGE Automated Hazard Detection in Construction Sites\\
Using Large Language and Vision-Language Models\par}

\vspace{2cm}

{\large \textbf{Author:} Islem Sahraoui\par}

\vspace{2cm}

{\large A thesis submitted for the degree of\par}

\vspace{0.5cm}

{\large Master of Science\par}

\vspace{0.5cm}

{\large at the\par}

\vspace{0.5cm}

{\large University of Houston\par}

\vspace{0.5cm}

{\large
Cullen College of Engineering\\
Department of Civil and Environmental Engineering\par}
\vspace{1cm}

{\large December 2025\par}
\end{titlepage}

\begin{abstract}
    \doublespacing
This thesis explores a multimodal AI framework for enhancing construction safety through the combined analysis of textual and visual data. In safety-critical environments such as construction sites, accident data often exists in multiple formats, such as written reports, inspection records, and site imagery, making it challenging to synthesize hazards using traditional approaches. To address this, this thesis proposed a multimodal AI framework that combines text and image analysis to assist in identifying safety hazards on construction sites. Two case studies were consucted to evaluate the capabilities of large language models (LLMs) and vision–language models (VLMs) for automated hazard identification.
The first case study introduces a hybrid pipeline that utilizes GPT 4o and GPT 4o mini to extract structured insights from a dataset of 28,000 OSHA accident reports (2000–2025). From this corpus, 100 reports were sampled to evaluate feasibility. The LLMs were used to summarize incident causes, injury types, and contributing factors. These outputs guided a VLM to detect and localize safety hazards, such as missing PPE or dangerous equipment in construction site images. The combined reasoning from text and image analysis enabled contextual safety assessments that mirror real-world site evaluations.
The second case study extends this investigation using Molmo 7B and Qwen2 VL 2B,  lightweight, open-source VLMs. Using the public ConstructionSite10k dataset, the performance of the two models was evaluated on rule-level safety violation detection using natural language prompts. This experiment served as a cost-aware benchmark against proprietary models and allowed testing at scale with ground-truth labels. Despite their smaller size, Molmo 7B and Quen2 VL 2B showed competitive performance in certain prompt configurations, reinforcing the feasibility of low-resource multimodal systems for rule-aware safety monitoring.
Together, these studies highlight both the opportunities and limitations of multimodal models in construction safety
applications. The results offer insight into model selection, cost–performance tradeoffs, and the path toward scalable, transparent safety intelligence systems.

\end{abstract}


\pagenumbering{arabic}

\section{Introduction} \label{Introduction}

The construction industry is widely recognized as one of the most hazardous industries with a high rate of occupational injuries and fatalities \cite{rabbi2024ai}. For example, 2023, about 1 in 5 workplace deaths occurred in construction (20.8\%), and 38.5\% of those were due to falls, slips, and trips \cite{BLS2022CFOI}. Despite the implementation of various safety regulations and practices, accidents persist due to the inherently dynamic and unpredictable nature of construction sites \cite{MAALI2024105231}. A key limitation of traditional construction safety management lies in its inability to efficiently collect, process, and act on diverse sources of safety information. Such traditional approaches typically rely on manual inspections and periodic risk assessments, which are often reactive and time-consuming \cite{MUSARAT2024102057}. Moreover, valuable historical safety data remains underutilized due to the difficulty of manually processing large volumes of unstructured accident reports \cite{10.1108/CI-04-2023-0062}. 
In the past decade, computer vision techniques have been used for detecting construction hazards and identifying compliance issues such as workers lacking PPE, improperly placed materials, and workers in restricted areas \cite{arshad2023computer,ren2021review, liu2021method}. Deep neural network models, particularly Convolutional Neural Network (CNN) such as You Only Look Once (YOLO) and region-based CNNs, have been widely used for object detection and segmentation in construction related research \cite{9650892}. 

Recent advances in Artificial Intelligence (AI) offer a path toward proactive, data-driven safety management \cite{gao2013verb-based,duggempudi2025text-to-layout:,chenchu2025signals}. Natural language processing techniques can extract information from unstructured safety text (e.g., OSHA narratives) \cite{gao2016public,ahmadi2025automatic}. Vision-Language Models (VLMs) extend this capability to images by combining visual perception with contextual reasoning, improving hazard identification beyond conventional object detection alone \cite{adil2025using}. However, adoption remains limited due to several challenges, including data scale and quality requirements, site variability, and usability issues for non-expert users \cite{abioye2021artificial}. In particular, VLMs require large, diverse, and domain-specific datasets to perform reliably, yet constructing high-quality image-text datasets tailored for construction is time-consuming, labor-intensive, and often constrained by privacy or access limitations. Additionally, the usability of such models poses a barrier, as most construction personnel lack the technical expertise needed to fine-tune, deploy, or even interact with complex AI systems. These limitations hinder the integration of VLMs into everyday safety workflows on job sites.

To address these limitations, this study proposes a novel, prompt-engineered multi-modal AI framework that utilizes LLMs and VLMs. The framework performs two pipelines. First, the LLM is prompted to analyze unstructured OSHA accident reports. It extracts structured fields (e.g., site location, injury severity, worker demographics) from reports and classifies accidents into predefined accident categories. These classifications are then used to quantify hazard risk through frequency statistics. In the second pipeline, VLMs are used to interpret construction site images and localize safety hazards by linking visible objects to hazard risks. This study is organized into two case studies. The first case study uses GPT-4o and GPT-4o-mini to process a sample of OSHA reports and assist a vision-language model in identifying and explaining visual hazards based on the extracted text. The second case study explores the capability of  lightweight, open-source models, Molmo 7B and Quen2 VL 2B, to detect rule-level safety violations directly from images. Using the publicly available ConstructionSite10k dataset, we benchmark Molmo and Quen2’s performance in a cost-effective and scalable manner, highlighting its potential for real-world safety monitoring under limited computing resources. Accordingly, this research aims to investigate whether pre-trained, general-purpose Vision-Language Models (VLMs), both commercial and open-source, can effectively address key construction safety tasks such as hazard identification, classification, and localization. By evaluating these models without any fine-tuning or retraining, the thesis seeks to understand their capabilities and limitations in performing real-world construction safety analysis and support informed decisions on their deployment in safety management systems.

\section{Literature Review}



The built environment, which includes roads, bridges, utilities, and buildings, forms the physical foundation of modern society. These systems enable daily mobility and economic activity, with pavements in particular supporting the continuous flow of people and goods and shaping user-perceived service quality \cite{wang2024comprehensive, lawrence1990built, gao2011performance, ccimen2021construction, gao2023deep, de2019reframing, gao2008robust, daniotti2020digital, gao2010network-level, anderson2015energy, gao2022missing, yu2023pavement, lebaku2024deep,gao2019impacts,vanegas2003road,gao2013an,dhatrak2020considering,gao2010optimal,gao2024considering,gao2007using,gao2011integrated,zhang2018a,vemuri2020pavement,qiao2016transportation,gao2019evaluation}. Construction, rehabilitation, and maintenance of built environment assets are labor-intensive processes that take place in dynamic, complex, and often hazardous site conditions. These environments are filled with heavy machinery, elevated structures, and constantly moving workers and materials, making safety management a persistent challenge \cite{sun2025simulation-based, webb2015schedule}.

\subsection{Safety Challenges in Construction}
Construction safety is a multifaceted challenge due to the dynamic and complex nature of construction sites, which are characterized by constantly changing conditions, diverse activities, and a multitude of potential hazards. The inherent variability in site conditions, including lighting, weather, and layout changes, exacerbates the difficulty of maintaining safety standards \cite{webb2015schedule, sun2025simulation-based}. Traditional safety monitoring approaches, which rely heavily on manual inspections by safety personnel, are often inadequate due to their time-consuming and costly nature, as well as their inability to provide continuous oversight \cite{samsami2024systematic}. The dynamic interplay of objects, equipment, and personnel in three-dimensional spaces frequently results in visual obstructions and occlusions, which pose significant challenges for both human observers and automated systems \cite{vukicevic2024systematic}.
One of the most pressing safety concerns in construction is the so-called "Fatal Four," which includes falls, struck-by incidents, caught-in/between incidents, and electrical incidents. These four categories account for approximately 70\% of all reported fatalities in the construction industry \cite{albert2020focus}. Falls alone are responsible for more than a third of these fatalities, often occurring from roofs, ladders, and scaffolding \cite{dong2013fatal}. Despite the implementation of safety programs like the Occupational Safety and Health Administration's (OSHA) Construction Fatal Four initiative, these hazards remain prevalent due to the complex and evolving nature of construction sites \cite{newaz2025critical}.
The challenge of hazard recognition is further compounded by the underreporting of safety hazards and the variance in hazard recognition across different trades. Studies have shown that a significant percentage of safety hazards remain unrecognized in construction environments, with estimates suggesting that up to 57\% of hazards may go unnoticed \cite{uddin2023leveraging}. This underrecognition is often attributed to the assumption that workers inherently possess the ability to identify hazards, an assumption that empirical evidence has repeatedly disproven \cite{jeelani2017construction}. Training programs, while beneficial, have not consistently resulted in improved hazard recognition, partly due to systemic weaknesses in their design and delivery \cite{uddin2020hazard}.
The presence of small objects, such as personal protective equipment (PPE), adds another layer of complexity to construction safety. These objects are often difficult to detect due to their size and the cluttered nature of construction sites \cite{wu2019automatic}. The effectiveness of object recognition systems is frequently compromised by occlusions and the limited visibility of small objects, which are common in the dense and congested workspaces typical of construction sites \cite{wang2020hardhat}. Despite advancements in computer vision and deep learning technologies, the scarcity of annotated data for occluded objects remains a critical issue, limiting the performance of these systems in real-world scenarios \cite{ahmed2023personal}.
Moreover, the dynamic and cluttered nature of construction sites poses significant challenges for real-time hazard detection systems. Variable lighting conditions, reflective materials, and the diverse colors and shapes of workers' attire can all adversely affect the performance of detection systems \cite{wu2025advancing}. The integration of advanced technologies, such as Vision-Language Models (VLMs), offers a promising avenue for improving hazard detection and risk assessment by enabling the analysis of semantic relationships among various elements in construction scenes \cite{chen2025tailored}. These models combine the visual understanding capabilities of computer vision with the contextual reasoning power of large-scale language models, allowing for more comprehensive and adaptive hazard identification processes \cite{wong2025enhancing}.

\subsection{Integration of AI in Construction Safety: Current Practices and Challenges}
The integration of Artificial Intelligence (AI) into construction safety practices has evolved significantly over the years, with various AI methodologies being employed to enhance safety management \cite{lebaku2025assessing}. Initially, the focus was on text-based analyses using traditional statistical methods to extract insights from accident reports and safety documentation. This foundational work laid the groundwork for data-driven insights that could improve construction safety \cite{tixier2016automated}. As technology advanced, the adoption of Natural Language Processing (NLP) and machine learning techniques allowed for the automated classification of large text datasets, providing consistent and efficient extraction of valuable insights from textual data \cite{khan2024exploring,wu2014analysis, gao2021a}. This automation has been crucial in understanding accident causes and implementing preventive measures, thereby reducing injuries and fatalities on construction sites \cite{luo2023convolutional}.
In parallel, computer vision technologies have been increasingly applied to construction safety, particularly for analyzing visual data. Early computer vision techniques facilitated the detection and classification of safety-related objects and incidents, albeit with limitations in accuracy and scalability \cite{kulinan2024advancing}. The introduction of deep learning, particularly Convolutional Neural Networks (CNNs), marked a significant improvement in the automation of safety equipment detection and hazardous condition identification \cite{chen2025real,gao2021detection}. These advancements have enabled real-time detection of unsafe behaviors and conditions, such as the improper use of Personal Protective Equipment (PPE) and proximity to heavy machinery, thereby enhancing on-site safety monitoring \cite{baek2024deep}.
The recent development of Vision-Language Models (VLMs) represents a significant leap forward in construction safety management. VLMs combine the visual understanding capabilities of computer vision models with the contextual reasoning power of Large Language Models (LLMs), allowing for comprehensive construction scene analysis. This integration facilitates the generation of descriptive outputs, reasoning about object relationships, and interpreting complex scenes based on both image content and textual knowledge \cite{wang2025integrated}. Such models have been proposed to automate safety inspections, offering a more adaptive hazard identification process that potentially improves workplace safety \cite{pu2024autorepo}.
Despite these advancements, the integration of AI in construction safety is not without challenges. One significant hurdle is the requirement for large, annotated datasets for training machine learning models, which can be resource-intensive to collect and maintain \cite{han2024utilizing}. Furthermore, the dynamic and complex nature of construction sites presents additional challenges in terms of data heterogeneity and the need for models that can adapt to varying site conditions \cite{kim2023real}. The reliance on large-scale, manually annotated datasets also limits the scalability and adaptability of dense captioning techniques, which are crucial for generating semantically rich descriptions of construction scenes \cite{liu2020manifesting}.
The integration of text and visual data analytics has become more pronounced, leveraging advanced NLP techniques and transformer-based models for deeper analysis. This approach enables comprehensive insights into construction safety by combining text, visual, and audio data, aiming for holistic safety monitoring systems \cite{li2025systematic}. However, the construction industry still faces significant data challenges, particularly in managing large, messy, and heterogeneous datasets that encompass both text and image data. This complexity necessitates the development of robust AI models capable of processing and analyzing diverse data types to provide actionable safety insights \cite{li2024data}.

\subsection{Multimodal Models: Features, Capabilities, and Distinctions}
Multimodal models, particularly Vision-Language Models (VLMs), represent a significant evolution in AI, offering capabilities that surpass traditional unimodal approaches. Unimodal models, which process a single type of data, such as text or images, often fall short in complex environments like construction sites, where understanding the interplay between various elements is crucial\cite{zhang2024vision}. In contrast, multimodal models integrate multiple data types, enabling a more comprehensive understanding of environments by aligning and processing both visual and textual information \cite{yin2024survey}.
VLMs, a subset of multimodal models, are designed to overcome the limitations of unimodal approaches by combining the strengths of visual encoders and Large Language Models (LLMs). These models are trained to align text and image data through both contrastive and generative methods, allowing them to understand and generate coherent outputs that reflect the semantic relationships within a scene. For instance, VLMs can infer implicit hazards and adapt to different site conditions without the need for extensive retraining, addressing the data challenge that traditional models face due to their reliance on predefined classes and captions \cite{minderer2022simple}.
The integration of vision and language in VLMs allows for advanced capabilities such as grounding and cross-modal reasoning. Grounding refers to the model's ability to connect visual elements with corresponding textual descriptions, which is essential for accurately interpreting complex construction environments. Cross-modal reasoning enables the model to draw inferences based on the combined analysis of visual and textual data, enhancing its ability to detect and assess hazards in dynamic and diverse settings \cite{chaudhary2025prompt}.
One of the primary advantages of VLMs is their ability to address the data challenge inherent in construction safety applications. Traditional computer vision models require large, annotated datasets, which are labor-intensive and costly to produce. These datasets often lack the diversity needed to generalize across different construction scenarios, leading to decreased performance in real-world applications. VLMs, however, leverage the extensive datasets used to train LLMs, allowing them to process and reason over both visual and textual data without the need for exhaustive retraining \cite{lee2025generative}.
Recent advancements in VLMs have demonstrated their potential in the construction industry. For example, Fan et al. \cite{fan2024ergochat} built the ErgoChat system, which utilizes a visual-language model to assess ergonomic risks for construction workers, showcasing the model's ability to interactively analyze and interpret complex visual data. Similarly, the application of GPT-4V for construction progress monitoring highlights the versatility of VLMs in tracking and analyzing development changes over time \cite{ersoz2024demystifying}.
Several powerful VLMs have emerged, each characterized by unique architectural innovations or training methodologies. These models have been made publicly available, facilitating their adoption and adaptation across various domains, including construction safety. The BLIP-2 framework, for instance, integrates visual encoders with LLMs to generate outputs that combine visual perception with linguistic comprehension, enabling a more nuanced understanding of construction environments \cite{li2023blip}.
The ability of VLMs to process and reason over multimodal data makes them particularly suited for addressing the challenges of construction safety. By understanding spatial relationships and inferring implicit hazards, VLMs can provide a more adaptable and context-aware approach to hazard detection. This capability is crucial in construction environments, where conditions are constantly evolving, and the ability to dynamically interpret diverse scenarios is essential for effective safety management \cite{li2022computer}.

\subsection{Using LLMs for Safety Documents and Text}
The integration of Large Language Models (LLMs) into the analysis of construction safety documents has opened new avenues for improving safety management practices. These models, such as GPT and LLaMA, have demonstrated significant potential in processing and understanding large volumes of unstructured text data, such as accident reports and safety narratives, which are critical for developing effective risk management strategies \cite{brown2020language, touvron2023llama}. The application of LLMs in this domain primarily revolves around tasks such as classification, information extraction, summarization, and compliance support.
One of the key applications of LLMs in construction safety is the classification of accident narratives. For example, \cite{ahmadi2025automatic} analyzed construction accident reports using LLMs such as GPT-3.5, GPT-4.0, Gemini Pro, and LLaMA 3.1. They classify incidents into different root causes, injury types, affected body parts, severities, and accident timing. Similarly, smetana et al.\cite{smetana5162763improving} also used GPT-3.5 to summarize and categorize over OSHA highway construction accident reports. Traditional methods, including support vector machines (SVMs) and convolutional neural networks (CNNs), have been employed to categorize accident causes with high precision \cite{qiao2022construction}. However, these methods often fall short in fully exploiting the capabilities of LLMs, which offer greater contextual awareness and the ability to process complex textual data more effectively \cite{ahmadi2025automatic}. For instance, a study employing a hybrid structured deep neural network with Word2Vec demonstrated improved semantic analysis of accident causes, enhancing classification accuracy by capturing nuanced relationships in accident data \cite{zhang2022hybrid}. 
LLMs have also been utilized for information extraction from  safety documents. In instance, sabetta et al. \cite{sabetta2025comparative} used GPT-4 Turbo to extract key attributes, such as the nature of injury, degree of fatality, and worker occupation from OSHA Lockout/Tagout accident narratives. The ability of LLMs to extract relevant and actionable safety information from complex documents, such as safety regulations and incident reports, marks a substantial leap forward in natural language processing \cite{tran2024leveraging}. This capability extends to extracting specific guidelines on machinery handling procedures, thereby enhancing the precision of safety compliance measures. Moreover, LLMs can be customized to project-specific datasets, allowing for the processing of proprietary documents and ensuring the confidentiality of sensitive project-related information.
In addition to classification and information extraction, LLMs have been employed for summarization tasks. For example, in highway construction safety analysis, LLMs were used to cluster, summarize, and identify causes and patterns across the OSHA Severe Injury Reports database, exhibiting high precision and recall in identifying key causes of accidents \cite{smetana2024highway}. This approach not only aids in evaluating resulting clusters but also eliminates the time consuming process of manually dissecting commonalities among incidents. Another example for summarization tasks by \cite{baek2025automated} developed a RAG-based framework combining a fine-tuned text retriever and a frozen LLM to generate safety guidance from over 64,000 accident cases. The model produced outputs comparable to expert-written reports based on BLEU, ROUGE, and BERTScore evaluations. similar to that, uhm et al.\cite{uhm2025effectiveness} tested RAG-GPT using LangChain to retrieve and summarize safety documents. Compared with standard GPT models, RAG-GPT achieved higher accuracy and contextual relevance in generating safety instructions, as confirmed by expert evaluations.. Another application of LLMs in the implementation with construction software in order to automate safety checking, \cite{madireddy2025large} integrated LLMs with BIM to automate building code compliance checks, which focuses on safety-related rules such as exit width, guardrails, and ventilation.

Despite the promising applications of LLMs in construction safety, several limitations persist. One significant challenge is the need for domain-specific tuning to address knowledge gaps in construction safety and accurately identify site-specific details \cite{kampelopoulos2025review}. Furthermore, the reliability of LLMs in safety critical applications is a concern, as inaccuracies or incomplete domain specifications can lead to ethical and liability issues. The quality of data used for training these models is also crucial, as inconsistent annotations or ambiguous reports can undermine performance \cite{shuang2024automatically}.

To address these limitations, prompt engineering and model fine tuning are essential strategies. Prompt engineering involves crafting clear instructions and contexts to guide LLMs in generating accurate outputs, while fine-tuning involves adjusting the model to better suit specific tasks or domains \cite{sammour2026responsible}. These strategies help mitigate the risks associated with deploying LLMs without understanding their strengths and limitations.

\subsection{Using VLMs for Construction Safety Images}
Vision-Language Models (VLMs) have emerged as a powerful tool for enhancing construction safety through image analysis, offering various applications such as object detection, image captioning, visual question answering (VQA), and scene understanding. These applications leverage the capabilities of VLMs to process and interpret visual data, thereby improving safety management on construction sites. For example, adil et al.\cite{adil2025using} proposed and validated a VLM-based framework to identify safety hazards from construction site images. The model processed safety guidelines and visual inputs to generate descriptive outputs of hazards, their severity, and mitigation suggestions. Similarly, tang et al.\cite{tang5179874double} introduced a double-thinking enabled VLM that enhances safety inspections by combining bounding box detection with chain-of-thought reasoning to reduce false positives and improve interpretability. Chen et al.\cite{chen2025vision} developed a VLM-powered framework to detect safety compliance issues, such as missing PPE, with fine-grained analysis and explainable annotations. \cite{chan2025context} proposed a context-aware VLM agent enriched with construction-specific ontology, integrating object detection outputs with domain reasoning to support site monitoring. Chauldhary at al.\cite{chaudhary2025prompt} proposed a comparative evaluation framework using multimodal LLMs such as GPT-4V, Gemini, Claude, and LLaVA to interpret construction site images and generate hazard. Chen et al.\cite{chen2025vision} fine-tuned a vision-language model on fire-safety inspection images to identify hazards and reference regulatory standards. The model improved BLEU and ROUGE scores, showing stronger accuracy and clarity than general multimodal LLMs. Xiao et al.\cite{xiao2024hazardvlm} trained a lightweight encoder-decoder model on nearly 8,000 hazard videos. Though not construction-specific, it showed a 9.6\% gain in accuracy and reduced false negatives by 27\%, demonstrating strong potential for real-time safety analysis.

Object detection is a critical application of VLMs in construction safety, primarily focusing on identifying personal protective equipment (PPE) compliance and detecting hazards. Models like YOLO (You Only Look Once) and its variants, such as YOLOv5, are widely used due to their speed and efficiency in real-time applications \cite{rabbi2024ai}. YOLOv5, for instance, has been employed to detect hardhat compliance and heavy equipment operation on construction sites, utilizing far-field surveillance videos to extract meaningful object information \cite{kim2023real}. The model's architecture, which includes modules like mosaic and spatial pyramid pooling, enhances its accuracy and processing speed, making it suitable for detecting small objects \cite{kim2023real}. However, challenges such as occlusion and the detection of small objects persist, necessitating further advancements in model capabilities.
In addition to object detection, image captioning and VQA are gaining traction as methods for providing contextual understanding of construction site images. These techniques allow for the generation of descriptive captions and the answering of queries related to visual content, thereby facilitating better safety assessments. The integration of natural language processing (NLP) with computer vision enhances this capability, enabling the extraction of insights from textual data such as safety regulations and incident reports \cite{mani2025leveraging}. This integration supports automated compliance assessment and improves hazard detection by linking visual information with regulatory guidelines \cite{adil2025using}.
Scene understanding, another application of VLMs, involves analyzing the spatial and temporal relationships between objects on construction sites. This is crucial for identifying potential hazards and ensuring worker safety. Techniques such as homography-based distance estimation and projective transformation are used to assess the proximity of workers to heavy equipment, thereby predicting collision risks \cite{kim2023real}. Furthermore, the use of deep learning models like Faster R-CNN has shown high accuracy in detecting workers and equipment, although the assignment of safety status to workers remains a challenge \cite{zhou2025integrating}.
Despite these advancements, several limitations and challenges remain in the application of VLMs for construction safety. The reliance on large, annotated datasets for training models poses a significant barrier, as these datasets are resource-intensive to create and may not generalize well across diverse construction scenarios \cite{ahmed2023personal}. Additionally, existing methods often fail to capture the semantic relationships within construction scenes, focusing instead on individual object detection without considering broader contextual interactions \cite{wu2021combining}. This lack of contextual understanding can hinder accurate hazard assessments, highlighting the need for more comprehensive approaches that integrate semantic reasoning with visual data analysis.
Moreover, the deployment of VLMs in construction safety is limited by factors such as cost, complexity, and the need for technical expertise, which smaller firms may lack \cite{chaudhary2025prompt}.

\subsection{Research Gaps and Limitations}
Despite significant potential, several barriers impede AI adoption and optimal performance. A pervasive challenge is the necessity of large, high-quality, extensively annotated datasets, both textual and visual. Insufficiently diverse data can cause weak generalization and overfitting, limiting real world applicability \cite{rabbi2024ai, gao2022missing, gao2019impacts}.
Technical challenges include computational demands of fine-tuning LLMs with heterogeneous data and managing reasoning flaws or hallucinations, where models produce plausible but inaccurate conclusions \cite{sammour2026responsible}. Additionally, generalization beyond training data remains problematic, as models often struggle with unfamiliar incident descriptions or varying textual styles \cite{rabbi2024ai}. 
Other critical challenge is the usability gap among non-expert users, such as construction workers and supervisors lacking deep AI technical expertise, which hampers widespread implementation \cite{mohy2024innovations}.
To address these gaps, this thesis adopts a prompt-based, training-free framework that leverages pre-trained LLMs and VLMs without the need for additional annotated datasets or fine-tuning. By evaluating both commercial and open-source models on real construction safety data, the study explores their practical usability and generalization potential under minimal technical setup.

\clearpage
\section{Methodology}
\begin{figure}[htbp]
    \centering
    \includegraphics[width=0.95\textwidth]{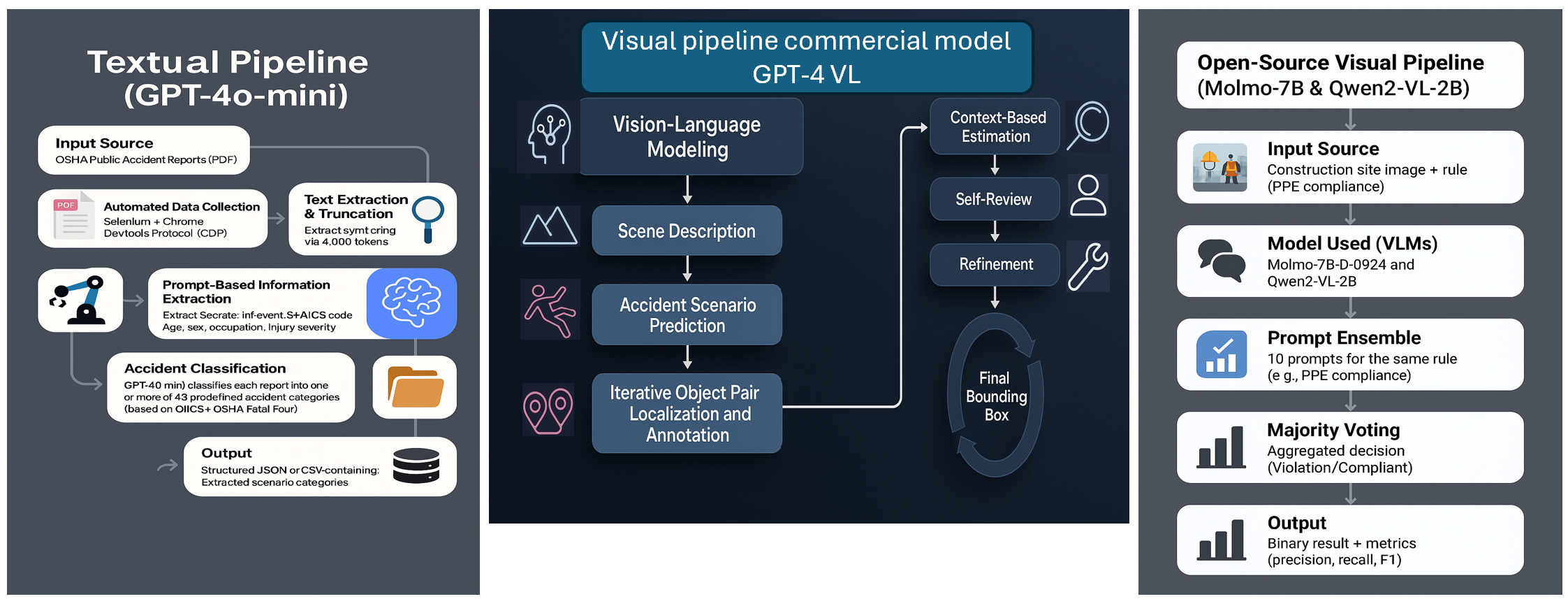}
    \caption{\parbox[t]{0.9\textwidth}{
    Overall methodology showing three parallel pipelines : (a) Textual pipeline using GPT-4o-mini, (b) Visual pipeline using GPT-4o Vision, and (c) Open-source visual pipeline using Molmo-7B and Qwen2-VL-2B.}}
    \label{fig:methodology_flowchart}
\end{figure}

This thesis presents two case studies focused on evaluating the potential of pre-trained multimodal AI models for construction safety assessment \cite{sahraoui2025integrating} \ref{fig:methodology_flowchart}. The primary case study introduces a two-pipeline framework that combines text and image analysis. The first pipeline processes textual data from OSHA accident reports, while the second pipeline analyzes construction site images to detect visual hazards. The primary implementation uses two models from the GPT-4o family. Specifically, GPT-4o-mini is employed for text analysis, and GPT-4o Vision is utilized for multimodal image understanding, both accessed via the OpenAI API. These models enable structured data extraction, accident scenario classification, and hazard detection through chained visual reasoning.

To complement the main framework, a secondary case study was conducted using Molmo 7B and Qwen2 VL 2B,  lightweight, open-source vision-language models. This experiment aimed to assess whether small, locally run VLMs could perform comparably to large proprietary models in construction safety tasks. The dataset and evaluation rule were adopted from the benchmark developed by chen et al.\cite{chen2025large} in their study on vision-language models for construction safety. Specifically, we focused on one safety rule related to PPE compliance and evaluated Molmo and Qwen's ability to reason about rule violations in construction site images. Both models were loaded and executed entirely within Google Colab, without the use of external APIs, providing a reproducible low-cost alternative. To enhance detection performance, over 20 prompt variations were iteratively tested and refined until achieving outputs that aligned with those produced by larger multimodal models.

All experiments were conducted in Google Colab using an NVIDIA T4 GPU. The textual pipeline used the GPT-4o model through OpenAI’s API. Each OSHA report was limited to 4,000 characters (around 1,000 tokens). Processing 100 reports requires roughly 12 minutes and costs about 1 USD in total API usage. The visual pipeline for image reasoning used GPT-4o Vision, and the rules violation detection used Molmo 7B D 0924, and Qwen2 VL 2B, all executed on the same T4 GPU. The open-source models (Molmo and Qwen2 VL 2B) were run locally using the Hugging Face Transformers library with automatic device mapping and mixed-precision inference (torch dtype='auto' for Molmo and 4 bit NF4 quantization for Qwen). Each image inference averaged 6–8 seconds per image with negligible cost since computations were performed locally.

\subsection{Textual Processing Pipeline}
The textual processing pipeline extracts structured data from OSHA inspection reports and analyzes incidents using an LLM. Among the available LLMs, the author tested different models and selected GPT-4o-mini because it provided the best balance between accuracy and cost. A structured prompt was used to guide GPT-4o-mini in extracting the needed information from each report. The details of this prompt and how it works are explained in the following sections.

\begin{figure}[htbp]
    \centering
    \includegraphics[width=0.85\textwidth]{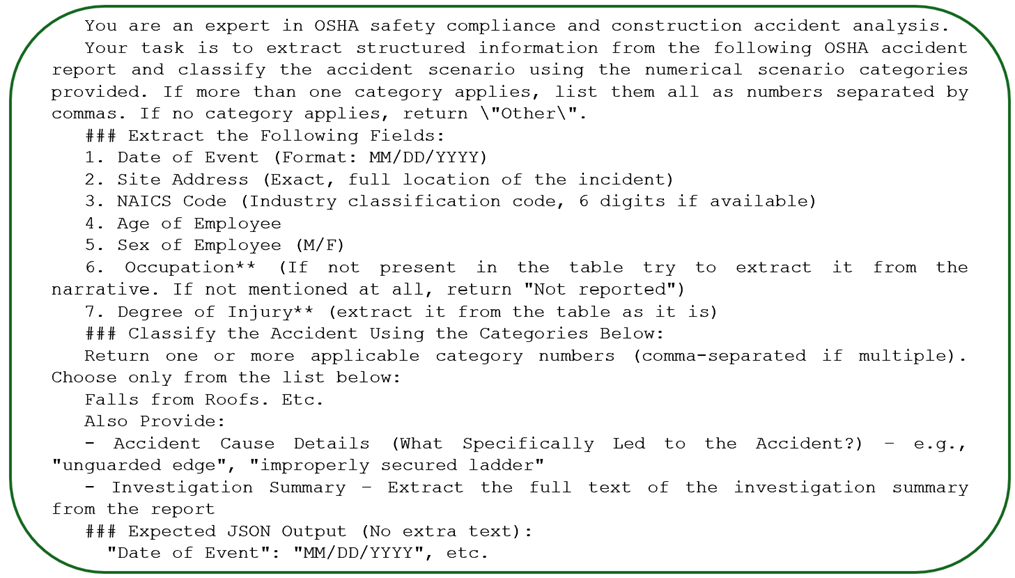}
    \caption{Pipeline prompt.}
    \label{fig:1}
\end{figure}

The first prompt in the framework Figures~\ref{fig:1} instructs the model to extract key fields, such as event date, location, occupation, and injury severity, and classify each report based on the predefined 43-category taxonomy Table~\ref{tab:incident_categories}. It is designed to return output in structured JSON format.

The taxonomy in Table 1 was developed using the Occupational Injury and Illness Classification System (OIICS, 1992) \cite{BLS1992OIICS} as a reference. The OIICS manual provides an extensive list of hazards, with over 100 construction-related hazards. To create a practical and manageable taxonomy, this extensive list was simplified by selecting the common and important accident types. Specifically, we identified 9 main categories and further divided these into 43 detailed subcategories. Our selection was guided by OSHA’s “Fatal Four” hazards (Falls, Struck-by, Caught-in/between, and Electrocutions) due to their frequent occurrence and severity in construction incidents. Moreover, we also considered OSHA Directorate of Training and Education’s (2011) “Top 10 Most Frequently Cited Standards” \cite{OSHA2011FocusFourFall} when creating this taxonomy.

\begin{table}[htbp]
\centering
\caption{Incident categories}
\renewcommand{\arraystretch}{1.4}
\setlength{\tabcolsep}{3pt}
\begin{tabular}{|>{\raggedright\arraybackslash}p{3 cm} | >{\raggedright\arraybackslash}p{12.5cm}|}
\hline
\textbf{Family} & \centering\textbf{Categories} \tabularnewline
\hline
\textbf{Fall} &
Falls from roofs, Falls from ladders, Falls from scaffolding, Falls through openings \& weak surfaces, Falls from elevated equipment \\
\hline
\textbf{Struck-by} &
Struck-by falling objects, Struck-by moving vehicles \& equipment, Struck-by swinging objects, Struck-by flying objects, Struck-by collapsing structures \\
\hline
\textbf{Caught-in/between} &
Trench \& excavation collapses, Caught-in machinery \& equipment, Caught between heavy equipment \& fixed objects, Structure collapse entrapment, Caught under overturned equipment \\
\hline
\textbf{Electrical} &
Contact with overhead power lines, Direct contact with energized electrical equipment, Electrical explosions \& arc flash, Improper grounding \& faulty wiring, Water \& electricity hazards \\
\hline
\textbf{Structure failures} &
Scaffold collapse, Building or roof collapse, Excavation or tunnel collapse, Bridge or crane failure \\
\hline
\textbf{Vehicle and heavy equipment accidents} &
Forklift \& equipment tip-overs, Cranes \& boom equipment failures, Vehicle backing accidents, Run-over by construction vehicles, Struck in work zone traffic \\
\hline
\textbf{Explosion and chemical exposure} &
Gas leaks leading to explosions, Flammable liquids or vapors igniting, Chemical spills \& hazardous material exposure, Oxygen deficiency in confined spaces, Hot work accidents (welding, cutting, grinding) \\
\hline
\textbf{Confined space and oxygen deficiency} &
Toxic gas exposure, Entrapment in confined spaces, Suffocation due to lack of oxygen, Explosion risks in confined spaces \\
\hline
\textbf{Environmental \& weather} &
Heat stress \& heat stroke, Cold stress \& hypothermia, Lightning strikes, Wind-related accidents \\
\hline
\end{tabular}
\label{tab:incident_categories}
\end{table}

\subsection{Visual Hazard Detection Pipeline}
\begin{figure}[htbp]
    \centering
    \includegraphics[width=0.85\textwidth]{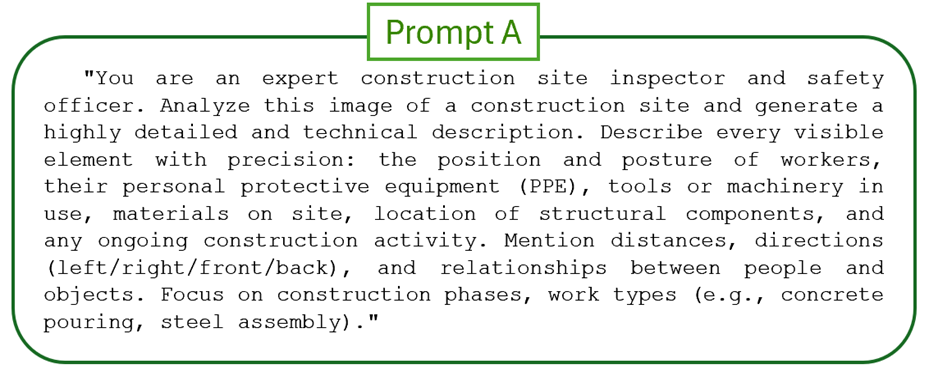}
    \caption{Scene description prompt.}
    \label{fig:2}
\end{figure}

The visual pipeline relies on a VLM, which combines visual understanding with natural language reasoning. In this pipeline, prompts were designed using the principle of Chain of Thought (CoT) prompting, where tasks are broken into intermediate reasoning steps \cite{wei2022chain}. Instead of requesting direct answers, the model was guided through successive stages of reasoning. Also an iterative self-review mechanism was used to improve accuracy. In this mechanism, initial bounding box predictions are reviewed and refined within cropped image segments. The first prompt in the visual reasoning Figures~\ref{fig:2} sequence instructs GPT-4o to generate a detailed and technical description of the given construction site image. This step establishes the semantic foundation for the CoT process and allows verification of the model’s situational understanding before hazard reasoning begins. 

\begin{figure}[htbp]
    \centering
    \includegraphics[width=0.85\textwidth]{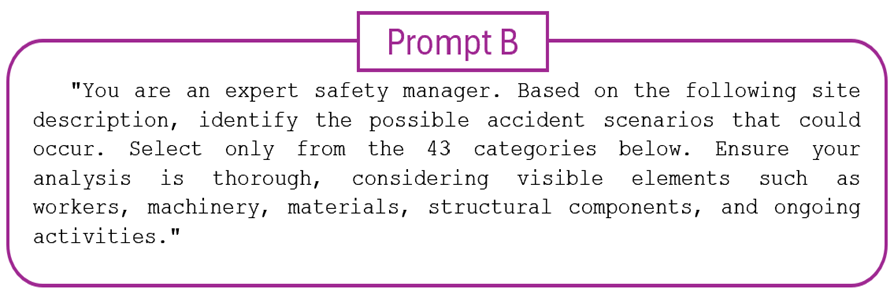}
    \caption{Accident Scenario Prediction Prompt.}
    \label{fig:3}
\end{figure}

The second prompt Figures~\ref{fig:3} builds directly on the previous scene description and instructs the model to infer possible accident scenarios. It uses conditional reasoning based on object interaction, spatial layout, and human activity. The logic mimics that of a human safety inspector: recognizing unsafe configurations or proximity violations.

\begin{figure}[htbp]
    \centering
    \includegraphics[width=0.85\textwidth]{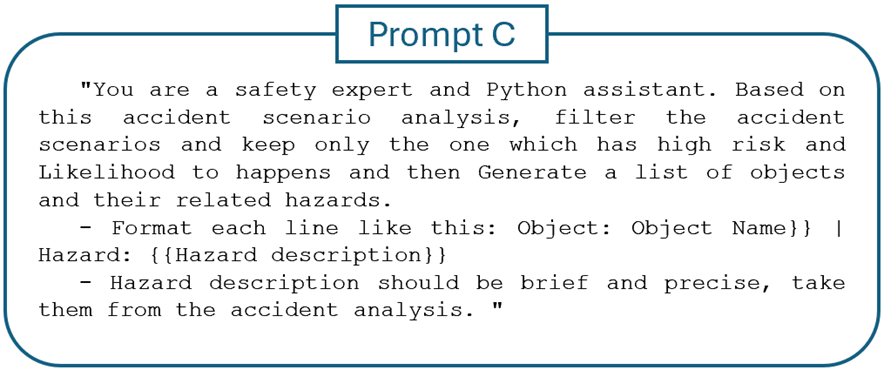}
    \caption{High-risk hazard filtering prompt.}
    \label{fig:4}
\end{figure}

The third prompt Figures~\ref{fig:4} refines the list of predicted hazards to focus on those with the highest risk and visual relevance. This step reduces output noise and ensures that only critical hazards are selected for visual annotation. In the prompt, filtering is guided by a predefined hazard taxonomy in Table 1.

\begin{figure}[htbp]
    \centering
    \includegraphics[width=0.85\textwidth]{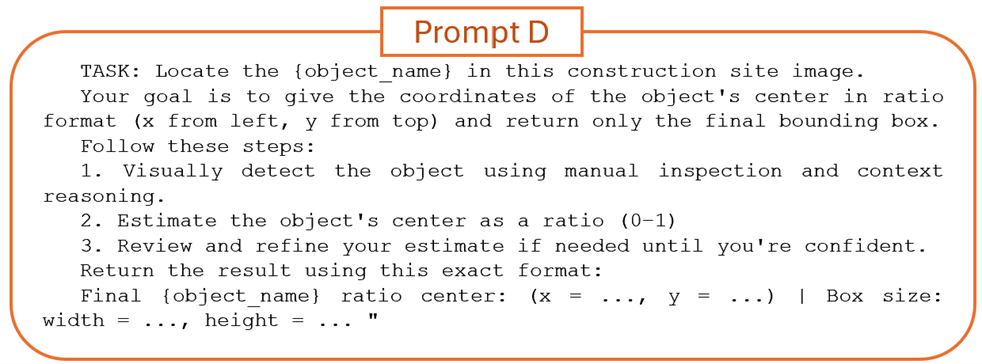}
    \caption{Object localization prompt.}
    \label{fig:5}
\end{figure}

The final prompt Figures~\ref{fig:5} in the pipeline guides the model to localize each high-risk object in the image and assign the appropriate hazard label. This prompt guides the model to estimate an object’s center in ratio format and define a bounding box. The process includes a self-review loop: the model is shown the initial detection, then asked to refine the box based on a cropped region. This two-stage prompt structure mimics human visual correction. It enhances detection accuracy by introducing error-aware refinement and feedback-based adjustment

\subsection{Multimodal Rule-Violation}
In the second part of this thesis, the capability of smaller, open‑source Vision‑Language Models (Molmo 7B and Qwen2 VL‑2B) was evaluated for construction safety rule‑checking. The aim of this analysis was to examine whether lightweight models can produce performance outcomes comparable to those of larger commercial VLMs when applied to rule‑based safety assessment tasks.

The author selected Molmo 7B and Quen2 VL 2B because they are open-source, lightweight, and cost-efficient model compared to large commercial VLMs. Among available small-scale models, Molmo and Qwen demonstrates relatively strong visual–language reasoning capabilities, making it well suited for evaluating construction safety QA while maintaining affordability and reproducibility.

The dataset and experimental setup was adopted from (Chen and Zou) paper, which includes labeled construction-site images and predefined safety rules. For this case study, the author focus on the rule “Use of basic PPE when on foot at construction sites.”

Because small VLMs are known to be sensitive to prompt phrasing, we conducted two separate evaluations. First, the author used a single short, clear, and precise prompt to assess the baseline performance of the model. Then, the evaluation was  extended using a prompt-ensemble approach, where each image was assessed using ten semantically equivalent prompts derived from the same safety rule. This allowed us to measure whether prompt diversity could enhance the model's robustness and consistency. The responses from the ten prompts were aggregated through majority voting to generate a single final decision per image, indicating whether a PPE violation was present.

To ensure consistent and interpretable responses from the vision-language model, ten prompts were carefully designed, these prompts were intentionally short, unambiguous, and focused, as shown in Table~\ref{tab:ppe-prompts-code}. Each prompt asked the same essential question, whether there was a violation of basic PPE requirements in the image, but with slightly different wordings or sentence structures. This variation enabled us to evaluate the model’s ability to generalize across semantically equivalent queries and mitigate prompt sensitivity, which is especially common in small models.

Unlike the original Chen and Zou setup, which used long, prompts that attempted to evaluate all safety rules at once, our approach emphasized simplicity, clarity and focus by assessing only one rule per evaluation. The author deliberately avoided complex sentence constructions or multi-step instructions to minimize confusion and reduce the risk of generating irrelevant or off-topic answers. 

Finally, the results obtained from the Molmo and Qwen models were compared with those reported in the original dataset paper to evaluate the performance difference between large and small multimodal models in construction safety image analysis.

\begin{table}[htbp]
\centering
\caption{Full list of PPE compliance prompts used for evaluation (Python format).}
\begin{tabular}{|c|p{12cm}|}
\hline
\textbf{Prompt ID} & \textbf{Prompt Text} \\
\hline
1 & In the image, is anyone not wearing basic PPE?\textbackslash n" + RULE\_LINE + "\textbackslash nAnswer only: Yes or No." \\
\hline
2 & Is anyone in this image violating basic PPE?\textbackslash n" + RULE\_LINE + "\textbackslash nAnswer only: Yes or No." \\
\hline
3 & Does the image show any person missing required PPE?\textbackslash n" + RULE\_LINE + "\textbackslash nAnswer one word: Yes or No." \\
\hline
4 & Is there at least one PPE wearing violation in the image?\textbackslash n" + RULE\_LINE + "\textbackslash nAnswer only: Yes or No." \\
\hline
5 & Are any people on foot not wearing basic PPE?\textbackslash n" + RULE\_LINE + "\textbackslash nAnswer only Yes or No." \\
\hline
6 & Check all visible people on the image for wearing the basic PPE.\textbackslash n" + RULE\_LINE + "\textbackslash nAny violation present?\textbackslash nAnswer only: Yes or No." \\
\hline
7 & Scan left-to-right: any person without basic PPE?\textbackslash n" + RULE\_LINE + "\textbackslash nAnswer only: Yes or No." \\
\hline
8 & Evaluate PPE compliance for people in the image.\textbackslash n" + RULE\_LINE + "\textbackslash nAny violation? \textbackslash nAnswer only: Yes or No." \\
\hline
9 & Check the image carefully and answer if anyone is not wearing basic PPE.\textbackslash n" + RULE\_LINE + "\textbackslash nIs there a violation? \textbackslash nAnswer only: Yes or No." \\
\hline
10 & Your task is to check whether any worker in the construction site is missing basic PPE.\textbackslash n" + RULE\_LINE + "\textbackslash nAnswer only: Yes or No." \\
\hline
\end{tabular}
\label{tab:ppe-prompts-code}
\end{table}

\section{Case study 1}
\subsection{Data Collection and Pre-processing}
\begin{figure}[H]
\centering
\includegraphics[width=0.85\textwidth]{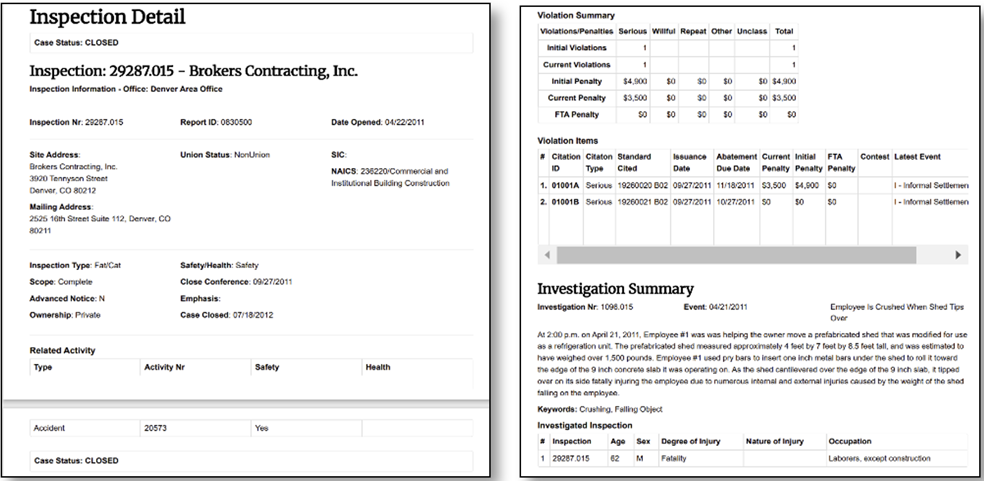}
\caption{Object localization prompt.}
\label{fig:6}
\end{figure}

In this case study, construction-related accident records from OSHA’s Integrated Management Information System (IMIS) were collected from January 2000 to March 2025. Using a headless Selenium WebDriver automation script, the system navigates OSHA’s online accident search interface and iteratively extracts links to over 28,000 construction-related accident reports dated between 2000 and 2025. For each record, the script follows the “Inspection Detail” link and generates a high-quality PDF version of the full report using the Chrome DevTools Protocol (Page.printToPDF). These reports are saved in a structured folder for further processing. To extract textual content, the pipeline uses the pdfplumber Python library, which parses each PDF and concatenates the text from all pages. If the document exceeds the token length allowed by the model, it is truncated to the first 4,000 characters. This preprocessing step ensures that critical content from the summary and narrative sections is retained. This automated Python pipeline was created to extract inspection report with key data such as ID, date, location, NAICS code, and summary Figures~\ref{fig:6}. A total of 28,575 individual inspection reports were collected. 

\subsection {Results of the Textual Processing Pipeline}
The output of the textual pipeline was a structured dataset in tabular format, with each row representing a processed OSHA inspection report as showen in table\ref{tab:osha_fields}. For each entry, the LLM model extracted critical fields such as event date, location, NAICS code, age, sex, occupation, and degree of injury. In addition, each incident was classified into one or more relevant accident scenario categories based on a predefined 43 categories. To evaluate the performance of the textual classification pipeline, a sample collected dataset was manually annotated by the author and compared with the LLM results. The model achieves an overall accuracy of 89\%.

\begin{table}[H]
\centering
\caption{Example of extracted fields and classified scenario from a sample OSHA accident report}
\label{tab:osha_fields}
\renewcommand{\arraystretch}{1.2}
\begin{tabular}{|p{4cm}|p{10cm}|}
\hline
\textbf{Field} & \textbf{Value} \\ \hline
Report ID & OSHA\_Inspection\_Report\_1433331.015 \\ \hline
Date of Event & 9/12/2019 \\ \hline
Site Address & 3170 SW Coral Way, Miami, FL 33135 \\ \hline
NAICS Code & 238290 \\ \hline
Age & 50 \\ \hline
Sex & M \\ \hline
Occupation & Assemblers \\ \hline
Degree of Injury & Hospitalized injury \\ \hline
Accident scenario & 7 \\ \hline
Accident Cause & Motor pack fell and crushed the employee \\ \hline
Investigation Summary & At 11:00 a.m. on September 12, 2019, an employee was dismantling a motor pack when it fell and crushed the employee, resulting in hospitalization. \\ \hline
\end{tabular}
\end{table}

\subsection{Visual Hazard Detection Pipeline Results}

The visual hazard pipeline was tested on 10 construction site images using GPT-4o Vision to evaluate its reasoning capabilities rather than object detection accuracy. The full step-wise process included scene description, accident scenario prediction, and hazard annotation. The goal was to support safety communication through contextual understanding rather than produce precise bounding boxes.

Each image underwent qualitative evaluation by a safety expert. For the narrative outputs (scene description and scenario prediction), the model’s ability to identify key visual elements such as workers, equipment, and activities, and infer appropriate accident scenarios was judged based on realism and relevance to actual site conditions. Despite the absence of formal ground truth, most outputs were coherent and matched common hazard interpretations.

For annotations, bounding boxes were checked visually. An annotation was accepted if it included the correct object and its label clearly communicated the associated hazard, even if the box was not perfectly aligned. Since preparing annotated datasets is time-consuming and GPT-4o usage is costly, testing was limited to 10 images. IoU metrics were not applied, as the goal was to help users visually identify risks rather than evaluate detection performance.

\begin{figure}[htbp]
    \centering
    \includegraphics[width=0.85\textwidth]{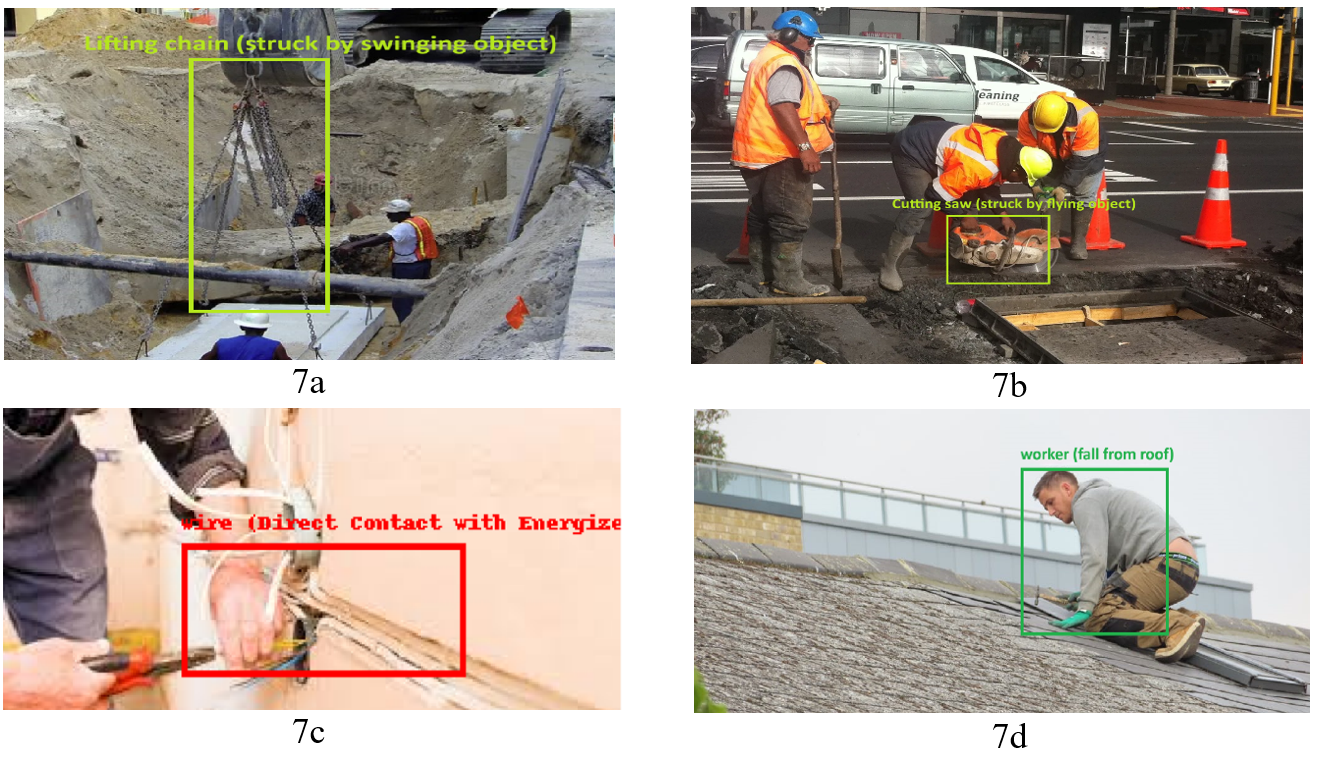}
    \caption{\parbox[t]{0.85\textwidth}{
    Final annotated outputs showing hazard localization across construction scenes using the visual pipeline:
    (a) Struck-by swinging object from a suspended lifting chain;
    (b) Struck-by flying object from a cutting saw;
    (c) Electrocution hazard from direct contact with energized wire;
    (d) Fall from roof due to missing fall protection.
    }}
    \label{fig:7}
\end{figure}

The results are shown in Figure \ref{fig:7}, which displays four annotated construction scenes representing three of OSHA’s major hazard families: Struck-by, Electrocution, and Falls. In Figure 7a, the model identified a suspended lifting chain during a trenching operation and labeled it as a struck-by swinging object hazard. It also recognizes both the chain’s movement and its proximity to nearby workers. Figure 7b shows road workers using a cutting saw, which was flagged as a struck-by flying object risk based on the tool type and active posture, which suggests potential for debris or tool malfunction. In Figure 7c, the model labeled exposed wiring being handled by a worker as direct contact with energized equipment. This accurately reflects an electrocution hazard through the positioning of the hand and wire. Finally, in Figure 7d, a worker kneeling on a sloped roof without visible fall protection was annotated as a fall from roof hazard. This shows the model’s ability to relate elevation, body posture, and missing safety gear to fall risk.

\clearpage
\section{Case study 2 : }
\subsection{Dataset collection}
\begin{figure}[H]
    \centering
    \includegraphics[width=0.85\textwidth]{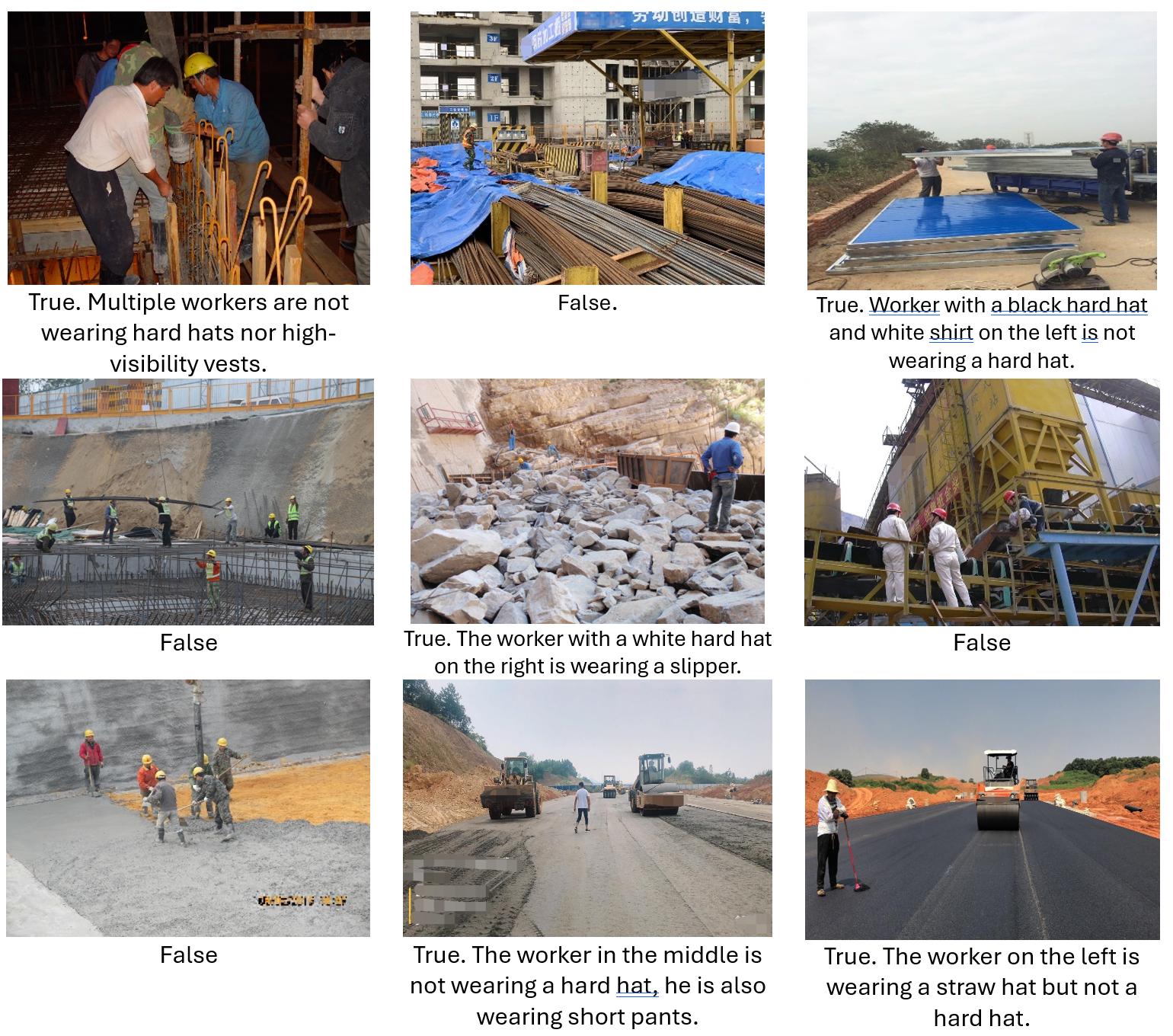}
    \caption{example of images with Q/A annotations.}
    \label{fig:9}
\end{figure}

The dataset used in this case study was adopted from Chen and Zou (2024). The authors introduced ConstructionSite-10K, a benchmark of around 10,000 annotated construction-site images labeled according to safety rules such as PPE use, fall protection, and hazard proximity. The dataset supports multiple evaluation modes, with the Rule-Based Visual Question Answering (Rule-VQA) task serving as the core component. This task requires the model to answer yes/no questions about compliance with specific safety rules. Our evaluation focused on this Rule-VQA setting, using the subset for Rule 1: “Use of basic PPE when on foot at construction sites” Figures~\ref{fig:9}, to test Molmo’s ability to identify rule violations through question–answer reasoning.

\subsection{Multimodal Rule-Violation results}
Using a prompt-ensemble approach (10 semantically equivalent prompts per image with majority voting), Molmo-7B achieved a Precision of 58.2\%, Recall of 79.6\% and F1 score of 67.2\%. In contrast, when tested with only a single prompt, the model achieved higher precision (85.7\%) but significantly lower recall (48.0\%), resulting in an F1 score of 61.5\%. This demonstrates that while single-prompt evaluation yields fewer false positives, it also misses many true violations. The prompt-ensemble method, by aggregating across multiple semantically equivalent queries, improved recall and achieved a more balanced overall performance.

Similarly, the Qwen2-VL-2B model obtained a Precision of 66.0\%, Recall of 67.3\%, and F1 score of 66.7 when using one prompt. When extended to 10 prompts, its performance improved to a Precision of 67.2\%, Recall of 98.0\%, and F1 score of 72.6\%, highlighting a consistent benefit of multi-prompt evaluation. This confirms that small vision-language models benefit substantially from semantic prompting and majority voting, which mitigate prompt sensitivity and stabilize responses.

When compared with the larger models reported in the Chen and Zou dataset paper, both Molmo and Qwen2-VL substantially outperform traditional baselines such as GPT (Precision 20.4\%, Recall 76.4\%, F1 32.2\%), GPT-5-shot (F1 30.2\%), and the LLaVA variants (F1 ranging from 13.6\% to 19.7\%). These results illustrate that lightweight, open-source VLMs, when paired with clear, single-rule prompts and semantic ensembles, can achieve performance comparable to or exceeding large commercial models on targeted safety rule-checking tasks. The key methodological difference lies in our focus on one rule per evaluation and the use of short, unambiguous prompts, whereas the original study used long, multi-rule prompts that introduced linguistic and contextual ambiguity.

\begin{table}[H]
\centering
\caption{\parbox[t]{0.85\textwidth}{
Performance comparison of vision-language models on construction safety detection (Rule 1). F1-scores computed from precision and recall, results (in \%).
}}
\begin{tabular}{|l|c|c|c|}
\hline
\textbf{Model} & \textbf{Precision} & \textbf{Recall} & \textbf{F1} \\
\hline
{GPT}               & 20.4  & 76.4  & 32.2 \\ {GPT 5-shot}        & 18.2  & 89.4  & 30.2 \\
{LLaVA 13B}         & 12.0  & 54.0  & 19.6 \\
{LLaVA 13B CoT}     & 12.0  & 55.0  & 19.7 \\
{LLaVA 34B 1-shot}  & 14.3  & 13.0  & 13.6 \\
{Qwen2-VL-2B}       & \textbf{66.0} & \textbf{67.3} & \textbf{66.7} \\
{Qwen2-VL-2B(10 prompt)}       & \textbf{67.2} & \textbf{98.0} & \textbf{72.6} \\
{Molmo-7B }         & \textbf{85.7} & \textbf{48.0} & \textbf{61.5} \\
{Molmo-7B (10 prompt) }         & \textbf{58.2} & \textbf{79.6} & \textbf{67.2} \\
\hline
\end{tabular}
\label{tab:model_comparison_f1}
\end{table}

\section{Discussion}
\subsection{Key Findings and Comparative Analysis}
This research investigated the potential of multimodal artificial intelligence frameworks to enhance construction safety by integrating textual and visual data analysis. Across the two case studies, the results demonstrate that prompt-based approaches using pre-trained Large Language Models (LLMs) and Vision–Language Models (VLMs) can achieve high interpretability and competitive performance without requiring any fine-tuning or specialized training datasets.

In the first case study, the textual processing pipeline powered by GPT-4o-mini achieved an overall accuracy of 89\% in classifying OSHA accident reports based on a 43-category taxonomy derived from the Occupational Injury and Illness Classification System (OIICS). This result highlights the model’s ability to extract structured information and categorize complex narrative reports without additional training or fine-tuning. Misclassifications mainly occurred in reports with vague descriptions or unusual formatting, confirming the pipeline’s sensitivity to structural inconsistencies.

In parallel, the visual reasoning pipeline used GPT-4o Vision to analyze a small set of construction site images. Through a structured sequence of prompts, including scene description, scenario prediction, hazard filtering, and self-review, the model identified and localized contextual hazards such as falls, struck-by, and electrocution events.The author evaluation confirmed that the outputs were coherent and aligned with real-world hazard interpretation. The iterative self-review mechanism improved interpretability by refining annotations based on cropped views.

The second case study examined whether smaller, open-source VLMs could match or approach the performance of larger commercial models for safety rule-checking. Using the ConstructionSite-10k dataset and focusing on a single rule (PPE compliance), Molmo-7B and Qwen2-VL-2B were tested using both single prompts and a 10-prompt ensemble approach. Results showed that Qwen2-VL-2B achieved an F1 score of 72.6\% with prompt ensembles, while Molmo-7B reached 67.2\%, both significantly outperforming earlier baselines like GPT-5-shot or LLaVA-13B. These results confirm that prompt clarity and prompt diversity (semantic rewording) can substantially improve performance even in small models.

Compared to previous work, this thesis extends the findings of \cite{adil2025using} by combining visual hazard identification and textual accident classification within a unified conceptual framework. While their study focused primarily on generating descriptive hazard explanations, the present thesis incorporates explicit object localization and structured information extraction from regulatory accident reports. This work also builds on the benchmark defined by \cite{chen2025large}, demonstrating that prompt-based strategies using lightweight, open‑source Vision–Language Models can achieve performance comparable to, or in some cases exceeding, larger fine‑tuned models for specific safety rule‑checking tasks. Furthermore, the results align with the observations of \cite{chaudhary2025prompt} regarding the effectiveness of concise, rule‑specific prompting and the performance instability observed when models are required to reason across multiple safety rules simultaneously.

\subsection{Limitations}
Despite promising results, several limitations were observed. First, the dataset size was constrained: only 100 OSHA reports and 10 construction images were analyzed for the for case study and 100 images for the second case study, which restricts statistical generalization. The absence of ground-truth bounding boxes prevented quantitative evaluation using metrics such as Intersection-over-Union (IoU). Second, the textual pipeline exhibited sensitivity to variations in report structure, occasionally failing to extract information from tables or irregular formats. Third, model outputs were highly dependent on prompt formulation, minor wording changes could alter reasoning quality or completeness. Finally, computational costs and API rate limits constrained larger-scale experimentation, and real-time performance was not evaluated under field conditions.

These limitations suggest that while the framework is effective for exploratory research and educational use, further work is required for operational deployment and large-scale validation.

\subsection{Practical Implications}

This framework offers a training-free, low-barrier AI solution for construction safety analysis. By combining GPT-based report classification and VLM-based hazard localization, safety officers and researchers can extract insights from unstructured narratives and assess risks in site images with minimal technical setup.

The successful use of small models like Qwen2-VL-2B and Molmo-7B shows that effective safety analysis can be achieved without large commercial systems. Semantic prompt ensembles further improve model stability, making the approach viable for practical use in compliance reviews, audits, or automated inspection tools.

In academic and training settings, the framework serves as a replicable case of prompt engineering and multimodal reasoning. It provides a foundation for future extensions, such as integration with BIM tools, live video analysis, or real-time feedback systems for field safety monitoring.

\section{Conclusions}
This study introduced a multi-modal AI framework that uses pre-trained large language and vision-language models to support construction safety analysis. The framework consists of two pipelines: a textual pipeline that extracts structured information and classifies OSHA accident reports, and a visual pipeline that identifies and localizes hazards in construction site images.

The textual pipeline, powered by GPT-4o-mini, achieved 89\% accuracy on a manually labeled test set. It successfully extracted key fields and classified incidents using a detailed 43-category taxonomy. However, the model showed sensitivity to structural variation in report formats.

The visual pipeline used GPT-4o vision and a sequence of prompts to reason through construction scenes. It correctly identified high-risk objects and visualized hazards using bounding boxes and annotations. The framework does not require fine-tuning, training data, or domain-specific customization. This makes it lightweight and accessible for safety research and practice. Limitations include sensitivity to input structure, dependence on prompt design, and the need for broader testing across larger and more diverse datasets.

To examine the role of model size and prompting technique, the study also evaluated smaller open‑source vision‑language models. These models demonstrated that, when supported by appropriate prompt formatting, lightweight architectures can achieve competitive performance in rule‑driven hazard assessment tasks.

More broadly, the findings indicate that prompting strategy has a meaningful effect on performance: larger models tend to benefit from more descriptive prompts, while smaller models respond more effectively to concise and focused instructions. This highlights the importance of aligning prompt complexity with model capacity rather than relying solely on model scale.

We observed that larger models such as GPT-4o are more responsive to detailed and context-rich prompts, whereas smaller models such as Qwen and Molmo benefit from short, precise, and clearly framed prompts. This underscores the importance of tailoring prompt complexity to model capacity.

Furthermore, semantic prompting, i.e., rephrasing queries in multiple ways while preserving meaning, proved effective in stabilizing model outputs and reducing sensitivity to wording variations. This approach, combined with prompt ensembles, offers a practical method for improving consistency and interpretability without requiring model fine-tuning.

Overall, the results show that prompt-based AI tools, when combined with the right prompting strategies, can support structured safety analysis using existing models and data. This offers a practical path for enhancing hazard identification in construction environments, even when using smaller and less resource-intensive models.

Future work may explore integrating real-time image capture, expanding evaluation across diverse site conditions, and combining textual and visual input into a unified multi-modal safety monitoring system. It is also important to expand the test samples, both textual and visual to better evaluate generalizability and performance across a wider range of accident types and site contexts. Further studies could assess the feasibility of deployment on edge devices or mobile safety tools.

\doublespacing
\footnotesize
\printbibliography[heading=bibintoc] 

@article{adil2025using,
  title={Using Vision Language Models for Safety Hazard Identification in Construction},
  author={Adil, Muhammad and Lee, Gaang and Gonzalez, Vicente A and Mei, Qipei},
  journal={arXiv preprint arXiv:2504.09083},
  year={2025}
}

@article{rabbi2024ai,
  title={AI integration in construction safety: Current state, challenges, and future opportunities in text, vision, and audio based applications},
  author={Rabbi, Ahmed Bin Kabir and Jeelani, Idris},
  journal={Automation in Construction},
  volume={164},
  pages={105443},
  year={2024},
  publisher={Elsevier}
}

@article{albert2020focus,
  title={Focus on the fatal-four: Implications for construction hazard recognition},
  author={Albert, Alex and Pandit, Bhavana and Patil, Yashwardhan},
  journal={Safety science},
  volume={128},
  pages={104774},
  year={2020},
  publisher={Elsevier}
}

@article{uddin2023leveraging,
  title={Leveraging ChatGPT to aid construction hazard recognition and support safety education and training},
  author={Uddin, SM Jamil and Albert, Alex and Ovid, Anto and Alsharef, Abdullah},
  journal={Sustainability},
  volume={15},
  number={9},
  pages={7121},
  year={2023},
  publisher={MDPI}
}

@article{khan2024exploring,
  title={Exploring associations between accident types and activities in construction using natural language processing},
  author={Khan, Numan and Nadeau, Sylvie and Pham, Xuan-Tan and Boton, Conrad},
  journal={Automation in Construction},
  volume={164},
  pages={105457},
  year={2024},
  publisher={Elsevier}
}

@article{luo2023convolutional,
  title={Convolutional neural network algorithm--based novel automatic text classification framework for construction accident reports},
  author={Luo, Xixi and Li, Xinchun and Song, Xuefeng and Liu, Quanlong},
  journal={Journal of Construction Engineering and Management},
  volume={149},
  number={12},
  pages={04023128},
  year={2023},
  publisher={American Society of Civil Engineers}
}

@article{arshad2023computer,
  title={Computer vision and IoT research landscape for health and safety management on construction sites},
  author={Arshad, Sameen and Akinade, Olugbenga and Bello, Sururah and Bilal, Muhammad},
  journal={Journal of Building Engineering},
  volume={76},
  pages={107049},
  year={2023},
  publisher={Elsevier}
}

@article{kulinan2024advancing,
  title={Advancing construction site workforce safety monitoring through BIM and computer vision integration},
  author={Kulinan, Almo Senja and Park, Minsoo and Aung, Pa Pa Win and Cha, Gichun and Park, Seunghee},
  journal={Automation in Construction},
  volume={158},
  pages={105227},
  year={2024},
  publisher={Elsevier}
}

@article{kim2023real,
  title={Real-time struck-by hazards detection system for small-and medium-sized construction sites based on computer vision using far-field surveillance videos},
  author={Kim, Hyung-soo and Seong, Jaehwan and Jung, Hyung-Jo},
  journal={Journal of Computing in Civil Engineering},
  volume={37},
  number={6},
  pages={04023028},
  year={2023},
  publisher={American Society of Civil Engineers}
}

@article{ahmadi2025automatic,
  title={Automatic construction accident report analysis using large language models (LLMs)},
  author={Ahmadi, Ehsan and Muley, Shashank and Wang, Chao},
  journal={Journal of Intelligent Construction},
  volume={3},
  number={1},
  pages={1--10},
  year={2025},
  publisher={TUP}
}

@article{chaudhary2025prompt,
  title={Prompt to Protection: A Comparative Study of Multimodal LLMs in Construction Hazard Recognition},
  author={Chaudhary, Nishi and Uddin, SM and Chandra, Sathvik Sharath and Ovid, Anto and Albert, Alex},
  journal={arXiv preprint arXiv:2506.07436},
  year={2025}
}

@article{lee2025generative,
  title={Generative AI-driven data augmentation for enhanced construction hazard detection},
  author={Lee, YeJun and Kang, GyeongNam and Kim, Jinwoo and Yoon, Seonghwan and Jeon, JungHo},
  journal={Automation in Construction},
  volume={177},
  pages={106317},
  year={2025},
  publisher={Elsevier}
}

@article{li2022computer,
  title={Computer vision-based hazard identification of construction site using visual relationship detection and ontology},
  author={Li, Yange and Wei, Han and Han, Zheng and Jiang, Nan and Wang, Weidong and Huang, Jianling},
  journal={Buildings},
  volume={12},
  number={6},
  pages={857},
  year={2022},
  publisher={MDPI}
}

@article{gao2008robust,
	 title = {Robust optimization for managing pavement maintenance and rehabilitation},
	 author = {Gao, L and Zhang, Z},
	 journal = {Transportation research record},
	 volume = {2084},
	 number = {1},
	 pages = {55-61},
	 year = {2008}
}

@article{gao2021a,
	 title = {A deep learning approach for imbalanced crash data in predicting highway-rail grade crossings accidents},
	 author = {Gao, L and Lu, P and Ren, Y},
	 journal = {Reliability Engineering \& System Safety},
	 volume = {216},
	 pages = {108019},
	 year = {2021}
}

@article{gao2016public,
	 title = {Public transit customer satisfaction dimensions discovery from online reviews},
	 author = {Gao, L and Yu, Y and Liang, W},
	 journal = {Urban Rail Transit},
	 volume = {2},
	 number = {3},
	 pages = {146-152},
	 year = {2016}
}

@article{gao2013verb-based,
	 title = {Verb-based text mining of road crash report},
	 author = {Gao, L and Wu, H},
	 journal = {TRB 92nd Annual Meeting},
	 year = {2013}
}

@article{wu2014analysis,
	 title = {Analysis of crash data using quantile regression for counts},
	 author = {Wu, H and Gao, L and Zhang, Z},
	 journal = {Journal of transportation engineering},
	 volume = {140},
	 number = {4},
	 pages = {04013025},
	 year = {2014}
}

@article{ren2021review,
	 title = {Review of emerging technologies and issues in rail and track inspection for local lines in the United States},
	 author = {Ren, Y and Lu, P and Ai, C and Gao, L and Qiu, S and Tolliver, D},
	 journal = {Journal of Transportation Engineering, Part A: Systems},
	 volume = {147},
	 number = {10},
	 pages = {04021062},
	 year = {2021}
}

@article{gao2021detection,
	 title = {Detection of pavement maintenance treatments using deep-learning network},
	 author = {Gao, L and Yu, Y and Ren, Y Hao and Lu, P},
	 journal = {Transportation Research Record},
	 volume = {2675},
	 number = {9},
	 pages = {1434-1443},
	 year = {2021}
}

@article{gao2010network-level,
	 title = {Network-level multi-objective optimal maintenance and rehabilitation scheduling},
	 author = {Gao, L and Xie, C and Zhang, Z},
	 journal = {Transportation Research Board 89th Annual MeetingTransportation Research Board},
	 year = {2010}
}

@article{gao2011performance,
	 title = {Performance modeling of infrastructure condition data with maintenance intervention},
	 author = {Gao, L and Aguiar-Moya, JP and Zhang, Z},
	 journal = {Transportation research record},
	 volume = {2225},
	 number = {1},
	 pages = {109-116},
	 year = {2011}
}

@article{webb2015schedule,
	 title = {Schedule compression impact on construction project safety},
	 author = {Webb, C and Gao, L and Song, L},
	 journal = {Front. Eng. Manag},
	 volume = {2},
	 pages = {344},
	 year = {2015}
}

@article{gao2013an,
	 title = {An augmented Lagrangian decomposition approach for infrastructure maintenance and rehabilitation decisions under budget uncertainty},
	 author = {Gao, L and Guo, R and Zhang, Z},
	 journal = {Structure and Infrastructure Engineering},
	 volume = {9},
	 number = {5},
	 pages = {448-457},
	 year = {2013}
}

@article{liu2021method,
	 title = {Method of evaluating and predicting traffic state of highway network based on deep learning},
	 author = {Liu, J and Wang, X and Li, Y and Kang, X and Gao, L},
	 journal = {Journal of Advanced Transportation},
	 volume = {2021},
	 number = {1},
	 pages = {8878494},
	 year = {2021}
}

@article{dhatrak2020considering,
	 title = {Considering deterioration propagation in transportation infrastructure maintenance planning},
	 author = {Dhatrak, O and Vemuri, V and Gao, L},
	 journal = {Journal of Traffic and Transportation Engineering (English Edition)},
	 year = {2020}
}

@article{gao2010optimal,
	 title = {Optimal infrastructure maintenance scheduling problem under budget uncertainty.},
	 author = {Gao, L and Zhang, Z},
	 journal = {Southwest Region University Transportation Center (US)},
	 year = {2010}
}

@article{gao2022missing,
	 title = {Missing pavement performance data imputation using graph neural networks},
	 author = {Gao, L and Yu, K and Lu, P},
	 journal = {Transportation research record},
	 volume = {2676},
	 number = {12},
	 pages = {409-419},
	 year = {2022}
}

@article{gao2024considering,
	 title = {Considering the spatial structure of the road network in pavement deterioration modeling},
	 author = {Gao, L and Yu, K and Lu, P},
	 journal = {Transportation Research Record},
	 volume = {2678},
	 number = {5},
	 pages = {153-161},
	 year = {2024}
}

@article{gao2019impacts,
	 title = {Impacts of seasonal and annual weather variations on network-level pavement performance},
	 author = {Gao, L and Hong, F and Ren, YH},
	 journal = {Infrastructures},
	 volume = {4},
	 number = {2},
	 pages = {27},
	 year = {2019}
}

@article{gao2007using,
	 title = {Using Markov process and method of moments for optimizing management strategies of pavement infrastructure},
	 author = {Gao, L and Zhang, Z and Tighe, SL},
	 journal = {Transportation Research Board 86th Annual MeetingTransportation Research Board},
	 year = {2007}
}

@article{gao2011integrated,
	 title = {Integrated maintenance and expansion planning for transportation network infrastructure},
	 author = {Gao, L and Xie, C and Zhang, Z and Waller, ST},
	 journal = {Transportation research record},
	 volume = {2225},
	 number = {1},
	 pages = {56-64},
	 year = {2011}
}

@article{gao2023deep,
	 title = {Deep learning–based pavement performance modeling using multiple distress indicators and road work history},
	 author = {Gao, L and Han, Z and Chen, Y},
	 journal = {Journal of Transportation Engineering, Part B: Pavements},
	 volume = {149},
	 number = {1},
	 pages = {04022061},
	 year = {2023}
}

@article{zhang2018a,
	 title = {A nested modelling approach to infrastructure performance characterisation},
	 author = {Zhang, Z and Gao, L},
	 journal = {International Journal of Pavement Engineering},
	 volume = {19},
	 number = {2},
	 pages = {174-180},
	 year = {2018}
}

@article{vemuri2020pavement,
	 title = {Pavement condition index estimation using smartphone based accelerometers for city of Houston},
	 author = {Vemuri, V and Ren, Y and Gao, L and Lu, P and Song, L},
	 journal = {Construction Research Congress},
	 volume = {2020},
	 pages = {522-531},
	 year = {2020}
}

@article{lebaku2024deep,
	 title = {Deep learning for pavement condition evaluation using satellite imagery},
	 author = {Lebaku, PKR and Gao, L and Lu, P and Sun, J},
	 journal = {Infrastructures},
	 volume = {9},
	 number = {9},
	 pages = {155},
	 year = {2024}
}

@article{madireddy2025large,
	 title = {Large Language Model-Driven Code Compliance Checking in Building Information Modeling},
	 author = {Madireddy, S and Gao, L and Din, ZU and Kim, K and Senouci, A and Han, Z and Zhang, Y},
	 journal = {Electronics},
	 volume = {14},
	 number = {11},
	 pages = {2146},
	 year = {2025}
}

@article{qiao2016transportation,
	 title = {Transportation and Economic Impact of Texas Short Line Railroads},
	 author = {Qiao, F and Gao, L and Saldarriaga, D and You, B and Li, Q and Song, L and Senouci, A and Dhatrak, O and ...},
	 journal = {Texas Department of Transportation},
	 year = {2016}
}

@article{vanegas2003road,
  title={Road map and principles for built environment sustainability},
  author={Vanegas, Jorge A},
  journal={Environmental science \& technology},
  volume={37},
  number={23},
  pages={5363--5372},
  year={2003},
  publisher={ACS Publications}
}

@article{anderson2015energy,
  title={Energy analysis of the built environment—A review and outlook},
  author={Anderson, John E and Wulfhorst, Gebhard and Lang, Werner},
  journal={Renewable and Sustainable Energy Reviews},
  volume={44},
  pages={149--158},
  year={2015},
  publisher={Elsevier}
}

@book{daniotti2020digital,
  title={Digital transformation of the design, construction and management processes of the built environment},
  author={Daniotti, Bruno and Gianinetto, Marco and Della Torre, Stefano},
  year={2020},
  publisher={Springer Nature}
}

@article{de2019reframing,
  title={Reframing construction within the built environment sector},
  author={de Valence, Gerard},
  journal={Engineering, Construction and Architectural Management},
  volume={26},
  number={5},
  pages={740--745},
  year={2019},
  publisher={Emerald Publishing Limited}
}

@article{ccimen2021construction,
  title={Construction and built environment in circular economy: A comprehensive literature review},
  author={{\c{C}}imen, {\"O}mer},
  journal={Journal of cleaner production},
  volume={305},
  pages={127180},
  year={2021},
  publisher={Elsevier}
}

@article{lawrence1990built,
  title={The built environment and spatial form},
  author={Lawrence, Denise L and Low, Setha M},
  journal={Annual review of anthropology},
  pages={453--505},
  year={1990},
  publisher={JSTOR}
}

@article{gao2019evaluation,
	 title = {Evaluation of transportation and economic impact of short line railroads in Texas},
	 author = {Gao, L and Saldarriaga, D and You, B and Qiao, F and Li, Q},
	 journal = {International Journal of Rail Transportation},
	 volume = {7},
	 number = {3},
	 pages = {191-207},
	 year = {2019}
}

@article{chenchu2025signals,
	 title = {Signals vs. Videos: Advancing Motion Intention Recognition for Human-Robot Collaboration in Construction},
	 author = {Chenchu, CG and Kim, K and Lu, G and Din, ZU},
	 journal = {arXiv preprint},
	 year = {2025}
}

@article{sahraoui2025integrating,
	 title = {Integrating generative ai in bim education: Insights from classroom implementation},
	 author = {Sahraoui, I and Kim, K and Gao, L and Din, Z and Senouci, A},
	 journal = {arXiv preprint},
	 year = {2025}
}

@article{lebaku2025assessing,
	 title = {Assessing the Influence of Pavement Performance on Road Safety Through Crash Frequency and Severity Analysis: PKR Lebaku et al.},
	 author = {Lebaku, PKR and Gao, L and Sun, J and Wang, X and Kang, X},
	 journal = {International Journal of Pavement Research and Technology,},
	 pages = {1-22},
	 year = {2025}
}

@article{sun2025simulation-based,
	 title = {Simulation-Based Framework for Predicting Construction Workforce Demand: A Comparative Analysis with Multivariate LSTM-Based Seq2Seq Model},
	 author = {Sun, J and Murphy, MR and Darren, H and Shuai, C and Gao, L},
	 journal = {Journal of Construction Engineering and Management},
	 year = {2025}
}

@article{wang2024comprehensive,
	 title = {Comprehensive network-level urban road asset valuation method integrating physical and social values},
	 author = {Wang, X and Li, Y and Zhang, R and Liu, J and Gao, L},
	 journal = {Journal of Transportation Engineering, Part A: Systems},
	 volume = {150},
	 number = {7},
	 pages = {04024024},
	 year = {2024}
}

@article{yu2023pavement,
	 title = {Pavement Missing Condition Data Imputation through Collective Learning-Based Graph Neural Networks},
	 author = {Yu, K and Gao, L},
	 journal = {International Conference on Transportation and Development},
	 volume = {2023},
	 pages = {416-423},
	 year = {2023}
}

@article{duggempudi2025text-to-layout:,
	 title = {Text-to-Layout: A Generative Workflow for Drafting Architectural Floor Plans Using LLMs},
	 author = {Duggempudi, J and Gao, L and Senouci, A and Han, Z and Zhang, Y},
	 journal = {arXiv preprint},
	 year = {2025}
}

@article{qiao2022construction,
  title={Construction-accident narrative classification using shallow and deep learning},
  author={Qiao, Jianfeng and Wang, Changfeng and Guan, Shuang and Shuran, Lv},
  journal={Journal of Construction Engineering and Management},
  volume={148},
  number={9},
  pages={04022088},
  year={2022},
  publisher={American Society of Civil Engineers}
}

@article{zhang2022hybrid,
  title={A hybrid structured deep neural network with Word2Vec for construction accident causes classification},
  author={Zhang, Fan},
  journal={International Journal of Construction Management},
  volume={22},
  number={6},
  pages={1120--1140},
  year={2022},
  publisher={Taylor \& Francis}
}

@article{tran2024leveraging,
  title={Leveraging large language models for enhanced construction safety regulation extraction},
  author={Tran, Si Van-Tien and Yang, Jaehun and Hussain, Rahat and Khan, Nasrullah and Kimito, Emmanuel Charles and Pedro, Akeem and Sotani, Mehrtash and Lee, Ung-Kyun and Park, Chansik},
  journal={Journal of Information Technology in Construction},
  volume={29},
  pages={1026--1038},
  year={2024},
  publisher={International Council for Research and Innovation in Building and Construction}
}

@article{smetana2024highway,
  title={Highway construction safety analysis using large language models},
  author={Smetana, Mason and Salles de Salles, Lucio and Sukharev, Igor and Khazanovich, Lev},
  journal={Applied Sciences},
  volume={14},
  number={4},
  pages={1352},
  year={2024},
  publisher={MDPI}
}

@article{kampelopoulos2025review,
  title={A review of LLMs and their applications in the architecture, engineering and construction industry},
  author={Kampelopoulos, Dimitrios and Tsanousa, Athina and Vrochidis, Stefanos and Kompatsiaris, Ioannis},
  journal={Artificial Intelligence Review},
  volume={58},
  number={8},
  pages={250},
  year={2025},
  publisher={Springer}
}

@article{shuang2024automatically,
  title={Automatically categorizing construction accident narratives using the deep-learning model with a class-imbalance treatment technique},
  author={Shuang, Qing and Liu, Xishan and Wang, Zhaojing and Xu, Xinxin},
  journal={Journal of Construction Engineering and Management},
  volume={150},
  number={9},
  pages={04024107},
  year={2024},
  publisher={American Society of Civil Engineers}
}

@article{sammour2026responsible,
  title={Responsible AI in construction safety: Systematic evaluation of large language models and prompt engineering},
  author={Sammour, Farouq and Xu, Jia and Wang, Xi and Hu, Mo and Zhang, Zhenyu},
  journal={Journal of Construction Engineering and Management},
  volume={152},
  number={1},
  pages={04025217},
  year={2026},
  publisher={American Society of Civil Engineers}
}

@article{wu2021combining,
  title={Combining computer vision with semantic reasoning for on-site safety management in construction},
  author={Wu, Haitao and Zhong, Botao and Li, Heng and Love, Peter and Pan, Xing and Zhao, Neng},
  journal={Journal of Building Engineering},
  volume={42},
  pages={103036},
  year={2021},
  publisher={Elsevier}
}

@article{mohy2024innovations,
  title={Innovations in safety management for construction sites: the role of deep learning and computer vision techniques},
  author={Mohy, Amr A and Bassioni, Hesham A and Elgendi, Elbadr O and Hassan, Tarek M},
  journal={Construction Innovation},
  year={2024},
  publisher={Emerald Publishing Limited}
}

@article{smetana5162763improving,
  title={Improving Large Language Model Assisted Categorization and Classification of Highway Construction Accidents},
  author={Smetana, Mason and Salles de Salles, Lucio and Khazanovich, Lev},
  journal={Available at SSRN 5162763}
}

@article{sabetta2025comparative,
  title={A comparative analysis for automated information extraction from OSHA Lockout/Tagout accident narratives with Large Language Model},
  author={Sabetta, Nicol{\`o} and Costantino, Francesco and Stabile, Sara},
  journal={Procedia Computer Science},
  volume={253},
  pages={1362--1372},
  year={2025},
  publisher={Elsevier}
}

@article{tang5179874double,
  title={A Double Thinking Enabled Visual Language Model for Open-Set Construction Site Safety Inspections},
  author={Tang, Yutong and Yan, Hui and Gao, Zeyu and Zhang, Zhen and Luo, Xiaochun},
  journal={Available at SSRN 5179874}
}

@article{chen2025vision,
  title={Vision language model for interpretable and fine-grained detection of safety compliance in diverse workplaces},
  author={Chen, Zhiling and Chen, Hanning and Imani, Mohsen and Chen, Ruimin and Imani, Farhad},
  journal={Expert Systems with Applications},
  volume={265},
  pages={125769},
  year={2025},
  publisher={Elsevier}
}

@article{chan2025context,
  title={Context-aware vision-language model agent enriched with domain-specific ontology for construction site safety monitoring},
  author={Chan, Chak-Fu and Wong, Peter Kok-Yiu and Guo, Xiaowen and Cheng, Jack CP and Chan, Jolly Pui-Ching and Leung, Pak-Him and Tao, Xingyu},
  journal={Automation in Construction},
  volume={177},
  pages={106305},
  year={2025},
  publisher={Elsevier}
}

@article{chen2025large,
  title={Are Large Pre-trained Vision Language Models Effective Construction Safety Inspectors?},
  author={Chen, Xuezheng and Zou, Zhengbo},
  journal={arXiv preprint arXiv:2508.11011},
  year={2025}
}

@article{baek2025automated,
  title={Automated safety risk management guidance enhanced by retrieval-augmented large language model},
  author={Baek, Seungwon and Park, Chan Young and Jung, Wooyong},
  journal={Automation in Construction},
  volume={176},
  pages={106255},
  year={2025},
  publisher={Elsevier}
}

@article{uhm2025effectiveness,
  title={Effectiveness of retrieval augmented generation-based large language models for generating construction safety information},
  author={Uhm, Miyoung and Kim, Jaehee and Ahn, Seungjun and Jeong, Hoyoung and Kim, Hongjo},
  journal={Automation in Construction},
  volume={170},
  pages={105926},
  year={2025},
  publisher={Elsevier}
}

@article{xiao2024hazardvlm,
  title={HazardVLM: A video language model for real-time hazard description in automated driving systems},
  author={Xiao, Dannier and Dianati, Mehrdad and Jennings, Paul and Woodman, Roger},
  journal={IEEE Transactions on Intelligent Vehicles},
  year={2024},
  publisher={IEEE}
}

@article{MAALI2024105231,
title = {Applications of existing and emerging construction safety technologies},
journal = {Automation in Construction},
volume = {158},
pages = {105231},
year = {2024},
issn = {0926-5805},
doi = {https://doi.org/10.1016/j.autcon.2023.105231},
url = {https://www.sciencedirect.com/science/article/pii/S0926580523004910},
author = {Omar Maali and Chien-Ho Ko and Phuong H.D. Nguyen},
}

@article{MUSARAT2024102057,
title = {Automated monitoring innovations for efficient and safe construction practices},
journal = {Results in Engineering},
volume = {22},
pages = {102057},
year = {2024},
issn = {2590-1230},
doi = {https://doi.org/10.1016/j.rineng.2024.102057},
url = {https://www.sciencedirect.com/science/article/pii/S2590123024003116},
author = {Muhammad Ali Musarat and Abdul Mateen Khan and Wesam Salah Alaloul and Noah Blas and Saba Ayub},
}

@article{10.1108/CI-04-2023-0062,
    author = {Mohy, Amr A. and Bassioni, Hesham A. and Elgendi, Elbadr O. and Hassan, Tarek M.},
    title = {Innovations in safety management for construction sites: the role of deep learning and computer vision techniques},
    journal = {Construction Innovation},
    year = {2024},
    month = {07},
    
    issn = {1471-4175},
    doi = {10.1108/CI-04-2023-0062},
    url = {https://doi.org/10.1108/CI-04-2023-0062},
    eprint = {https://www.emerald.com/ci/article-pdf/doi/10.1108/CI-04-2023-0062/9753014/ci-04-2023-0062.pdf},
}

@ARTICLE{9650892,
  author={Shanti, Mohammad Z. and Cho, Chung-Suk and Byon, Young-Ji and Yeun, Chan Yeob and Kim, Tae-Yeon and Kim, Song-Kyoo and Altunaiji, Ahmed},
  journal={IEEE Access}, 
  title={A Novel Implementation of an AI-Based Smart Construction Safety Inspection Protocol in the UAE}, 
  year={2021},
  volume={9},
  number={},
  pages={166603-166616},
  doi={10.1109/ACCESS.2021.3135662}
}

@article{abioye2021artificial,
  title={Artificial intelligence in the construction industry: A review of present status, opportunities and future challenges},
  author={Abioye, Sofiat O and Oyedele, Lukumon O and Akanbi, Lukman and Ajayi, Anuoluwapo and Delgado, Juan Manuel Davila and Bilal, Muhammad and Akinade, Olugbenga O and Ahmed, Ashraf},
  journal={Journal of Building Engineering},
  volume={44},
  pages={103299},
  year={2021},
  publisher={Elsevier}
}

@article{samsami2024systematic,
  title={A Systematic Review of Automated Construction Inspection and Progress Monitoring (ACIPM): Applications, Challenges, and Future Directions},
  author={Samsami, Reihaneh},
  journal={CivilEng},
  volume={5},
  number={1},
  pages={265--287},
  year={2024},
  publisher={MDPI}
}

@article{vukicevic2024systematic,
  title={A systematic review of computer vision-based personal protective equipment compliance in industry practice: advancements, challenges and future directions},
  author={Vukicevic, Arso M and Petrovic, Milos and Milosevic, Pavle and Peulic, Aleksandar and Jovanovic, Kosta and Novakovic, Aleksandar},
  journal={Artificial Intelligence Review},
  volume={57},
  number={12},
  pages={319},
  year={2024},
  publisher={Springer}
}

@article{dong2013fatal,
  title={Fatal falls from roofs among US construction workers},
  author={Dong, Xiuwen Sue and Choi, Sang D and Borchardt, James G and Wang, Xuanwen and Largay, Julie A},
  journal={Journal of Safety Research},
  volume={44},
  pages={17--24},
  year={2013},
  publisher={Elsevier}
}

@article{newaz2025critical,
  title={A critical analysis of construction incident trends and strategic interventions for enhancing safety},
  author={Newaz, Mohammad Tanvi and Jefferies, Marcus and Ershadi, Mahmoud},
  journal={Safety Science},
  volume={187},
  pages={106865},
  year={2025},
  publisher={Elsevier}
}

@article{jeelani2017construction,
  title={Why do construction hazards remain unrecognized at the work interface?},
  author={Jeelani, Idris and Albert, Alex and Gambatese, John A},
  journal={Journal of construction engineering and management},
  volume={143},
  number={5},
  pages={04016128},
  year={2017},
  publisher={American Society of Civil Engineers}
}

@article{uddin2020hazard,
  title={Hazard recognition patterns demonstrated by construction workers},
  author={Uddin, SM Jamil and Albert, Alex and Alsharef, Abdullah and Pandit, Bhavana and Patil, Yashwardhan and Nnaji, Chukwuma},
  journal={International Journal of Environmental Research and Public Health},
  volume={17},
  number={21},
  pages={7788},
  year={2020},
  publisher={MDPI}
}

@article{wu2019automatic,
  title={Automatic detection of hardhats worn by construction personnel: A deep learning approach and benchmark dataset},
  author={Wu, Jixiu and Cai, Nian and Chen, Wenjie and Wang, Huiheng and Wang, Guotian},
  journal={Automation in construction},
  volume={106},
  pages={102894},
  year={2019},
  publisher={Elsevier}
}

@article{wang2020hardhat,
  title={Hardhat-wearing detection based on a lightweight convolutional neural network with multi-scale features and a top-down module},
  author={Wang, Lu and Xie, Liangbin and Yang, Peiyu and Deng, Qingxu and Du, Shuo and Xu, Lisheng},
  journal={Sensors},
  volume={20},
  number={7},
  pages={1868},
  year={2020},
  publisher={MDPI}
}

@article{ahmed2023personal,
  title={Personal protective equipment detection: A deep-learning-based sustainable approach},
  author={Ahmed, Mohammed Imran Basheer and Saraireh, Linah and Rahman, Atta and Al-Qarawi, Seba and Mhran, Afnan and Al-Jalaoud, Joud and Al-Mudaifer, Danah and Al-Haidar, Fayrouz and AlKhulaifi, Dania and Youldash, Mustafa and others},
  journal={Sustainability},
  volume={15},
  number={18},
  pages={13990},
  year={2023},
  publisher={MDPI}
}

@article{wu2025advancing,
  title={Advancing construction safety: YOLOv8-CGS helmet detection model},
  author={Wu, Zhihui and Lei, Xiaojia and Kumar, Munish},
  journal={PloS one},
  volume={20},
  number={5},
  pages={e0321713},
  year={2025},
  publisher={Public Library of Science San Francisco, CA USA}
}

@article{chen2025tailored,
  title={Tailored vision-language framework for automated hazard identification and report generation in construction sites},
  author={Chen, Qihua and Yin, Xianfei},
  journal={Advanced Engineering Informatics},
  volume={66},
  pages={103478},
  year={2025},
  publisher={Elsevier}
}

@article{wong2025enhancing,
  title={Enhancing visual-LLM for construction site safety compliance via prompt engineering and Bi-stage retrieval-augmented generation},
  author={Wong, Peter Kok-Yiu and Cheng, Jack CP and Chan, Chak-Fu and Leung, Pak-Him and Tao, Xingyu and others},
  journal={Automation in Construction},
  volume={179},
  pages={106490},
  year={2025},
  publisher={Elsevier}
}

@article{tixier2016automated,
  title={Automated content analysis for construction safety: A natural language processing system to extract precursors and outcomes from unstructured injury reports},
  author={Tixier, Antoine J-P and Hallowell, Matthew R and Rajagopalan, Balaji and Bowman, Dean},
  journal={Automation in Construction},
  volume={62},
  pages={45--56},
  year={2016},
  publisher={Elsevier}
}

@article{chen2025real,
  title={Real-Time Detection of Personal Protective Equipment Violations for Construction Workers Using Semisupervised Learning and Video Clips},
  author={Chen, Qihua and Long, Danbing and Wang, Siqi and Chen, Qirong and Yuan, Beifei},
  journal={Journal of Construction Engineering and Management},
  volume={151},
  number={3},
  pages={04024213},
  year={2025},
  publisher={American Society of Civil Engineers}
}

@article{baek2024deep,
  title={Deep learning-based automated productivity monitoring for on-site module installation in off-site construction},
  author={Baek, Jongyeon and Kim, Daeho and Choi, Byungjoo},
  journal={Developments in the Built Environment},
  volume={18},
  pages={100382},
  year={2024},
  publisher={Elsevier}
}

@article{wang2025integrated,
  title={An integrated approach for automatic safety inspection in construction: Domain knowledge with multimodal large language model},
  author={Wang, Yiheng and Luo, Hanbin and Fang, Weili},
  journal={Advanced Engineering Informatics},
  volume={65},
  pages={103246},
  year={2025},
  publisher={Elsevier}
}

@article{pu2024autorepo,
  title={AutoRepo: A general framework for multimodal LLM-based automated construction reporting},
  author={Pu, Hongxu and Yang, Xincong and Li, Jing and Guo, Runhao},
  journal={Expert Systems with Applications},
  volume={255},
  pages={124601},
  year={2024},
  publisher={Elsevier}
}

@article{han2024utilizing,
  title={Utilizing synthetic images to enhance the automated recognition of small-sized construction tools},
  author={Han, Soeun and Park, Wonjun and Jeong, Kyumin and Hong, Taehoon and Koo, Choongwan},
  journal={Automation in Construction},
  volume={163},
  pages={105415},
  year={2024},
  publisher={Elsevier}
}

@article{liu2020manifesting,
  title={Manifesting construction activity scenes via image captioning},
  author={Liu, Huan and Wang, Guangbin and Huang, Ting and He, Ping and Skitmore, Martin and Luo, Xiaochun},
  journal={Automation in Construction},
  volume={119},
  pages={103334},
  year={2020},
  publisher={Elsevier}
}

@article{li2025systematic,
  title={A systematic review of multi-modal large language models on domain-specific applications},
  author={Li, Sirui and Wong, Kok Wai and Wang, Guanjin and Duong, Thach-Thao},
  journal={Artificial Intelligence Review},
  volume={58},
  number={12},
  pages={1--47},
  year={2025},
  publisher={Springer}
}

@article{li2024data,
  title={Data issues in industrial ai system: A meta-review and research strategy},
  author={Li, Xuejiao and Yang, Cheng and M{\o}ller, Charles and Lee, Jay},
  journal={arXiv preprint arXiv:2406.15784},
  year={2024}
}

@article{fan2024ergochat,
  title={Ergochat: a visual query system for the ergonomic risk assessment of construction workers},
  author={Fan, Chao and Mei, Qipei and Wang, Xiaonan and Li, Xinming},
  journal={arXiv preprint arXiv:2412.19954},
  year={2024}
}

@article{ersoz2024demystifying,
  title={Demystifying the Potential of ChatGPT-4 Vision for Construction Progress Monitoring},
  author={Ersoz, Ahmet Bahaddin},
  journal={arXiv preprint arXiv:2412.16108},
  year={2024}
}

@inproceedings{li2023blip,
  title={Blip-2: Bootstrapping language-image pre-training with frozen image encoders and large language models},
  author={Li, Junnan and Li, Dongxu and Savarese, Silvio and Hoi, Steven},
  booktitle={International conference on machine learning},
  pages={19730--19742},
  year={2023},
  organization={PMLR}
}

@inproceedings{minderer2022simple,
  title={Simple open-vocabulary object detection},
  author={Minderer, Matthias and Gritsenko, Alexey and Stone, Austin and Neumann, Maxim and Weissenborn, Dirk and Dosovitskiy, Alexey and Mahendran, Aravindh and Arnab, Anurag and Dehghani, Mostafa and Shen, Zhuoran and others},
  booktitle={European conference on computer vision},
  pages={728--755},
  year={2022},
  organization={Springer}
}

@article{brown2020language,
  title={Language models are few-shot learners},
  author={Brown, Tom and Mann, Benjamin and Ryder, Nick and Subbiah, Melanie and Kaplan, Jared D and Dhariwal, Prafulla and Neelakantan, Arvind and Shyam, Pranav and Sastry, Girish and Askell, Amanda and others},
  journal={Advances in neural information processing systems},
  volume={33},
  pages={1877--1901},
  year={2020}
}

@article{touvron2023llama,
  title={Llama: Open and efficient foundation language models},
  author={Touvron, Hugo and Lavril, Thibaut and Izacard, Gautier and Martinet, Xavier and Lachaux, Marie-Anne and Lacroix, Timoth{\'e}e and Rozi{\`e}re, Baptiste and Goyal, Naman and Hambro, Eric and Azhar, Faisal and others},
  journal={arXiv preprint arXiv:2302.13971},
  year={2023}
}

@article{mani2025leveraging,
  title={LEVERAGING LARGE LANGUAGE MODELS TO ENHANCE SAFETY AWARENESS AND ACCESSIBILITY OF OSHA REGULATIONS FOR CONSTRUCTION WORKERS},
  author={Mani, Nirajan and Pradhananga, Nipesh and Shrestha, Kishor},
  year={2025}
}

@article{zhou2025integrating,
  title={Integrating ontology and computer vision for intelligent monitoring of unsafe conditions in hot work},
  author={Zhou, Zhengwen and Chen, Shan and Kou, Junhui and Chen, Siqi and Liu, Jiaxin and Guo, Liangjie},
  journal={Automation in Construction},
  volume={180},
  pages={106574},
  year={2025},
  publisher={Elsevier}
}

@article{yin2024survey,
  title={A survey on multimodal large language models},
  author={Yin, Shukang and Fu, Chaoyou and Zhao, Sirui and Li, Ke and Sun, Xing and Xu, Tong and Chen, Enhong},
  journal={National Science Review},
  volume={11},
  number={12},
  pages={nwae403},
  year={2024},
  publisher={Oxford University Press}
}

@article{zhang2024vision,
  title={Vision-language models for vision tasks: A survey},
  author={Zhang, Jingyi and Huang, Jiaxing and Jin, Sheng and Lu, Shijian},
  journal={IEEE transactions on pattern analysis and machine intelligence},
  volume={46},
  number={8},
  pages={5625--5644},
  year={2024},
  publisher={IEEE}
}

@techreport{BLS1992OIICS,
  author       = {{U.S. Bureau of Labor Statistics}},
  title        = {Occupational Injury and Illness Classification Manual},
  institution  = {U.S. Department of Labor},
  address      = {Washington, DC},
  year         = {1992},
  url          = {https://www.bls.gov/iif/oiics_manual_2010.pdf},
  note         = {Original edition 1992; see later revisions as applicable}
}

@techreport{OSHA2011FocusFourFall,
  author       = {{OSHA Directorate of Training and Education}},
  title        = {Construction Focus Four: Fall Hazards (Instructor Guide)},
  institution  = {U.S. Department of Labor},
  address      = {Washington, DC},
  year         = {2011},
  month        = sep,
  url          = {https://www.osha.gov/sites/default/files/falls_ig.pdf},
  urldate      = {2025-11-10}
}

@article{wei2022chain,
  title={Chain-of-thought prompting elicits reasoning in large language models},
  author={Wei, Jason and Wang, Xuezhi and Schuurmans, Dale and Bosma, Maarten and Xia, Fei and Chi, Ed and Le, Quoc V and Zhou, Denny and others},
  journal={Advances in neural information processing systems},
  volume={35},
  pages={24824--24837},
  year={2022}
}

@techreport{BLS2022CFOI,
  author       = {{U.S. Bureau of Labor Statistics}},
  title        = {Census of Fatal Occupational Injuries (CFOI) — 2022 Summary},
  institution  = {U.S. Department of Labor},
  address      = {Washington, DC},
  year         = {2023},
  url          = {https://www.bls.gov/iif/},
  note         = {Tables include construction share of fatalities and falls/slips/trips share}
}


\end{document}